\documentclass[a4paper,fleqn]{cas-sc}

\usepackage[numbers,sort&compress]{natbib}
\def\tsc#1{\csdef{#1}{\textsc{\lowercase{#1}}\xspace}}
\tsc{WGM}
\tsc{QE}
\usepackage{booktabs, tabularx}
\usepackage{multirow}
\usepackage{threeparttable}
\usepackage{longtable}
\usepackage{threeparttablex}
\usepackage{enumitem}
\usepackage{amsmath,amsfonts,amssymb,accents}

\usepackage{algorithmic}
\usepackage{algorithm}
\usepackage{array}
\usepackage{graphicx}

\usepackage{rotating}
\usepackage[table]{xcolor}

\usepackage{hyperref}
\usepackage{adjustbox}
\usepackage{multicol}
\usepackage{color,soul}
\usepackage{caption}
\usepackage{float}

\makeatletter
\def\widebreve{\mathpalette\wide@breve}
\def\wide@breve#1#2{\sbox\z@{$#1#2$}%
     \mathop{\vbox{\m@th\ialign{##\crcr
\kern0.08em\brevefill#1{0.8\wd\z@}\crcr\noalign{\nointerlineskip}%
                    $\hss#1#2\hss$\crcr}}}\limits}
\def\brevefill#1#2{$\m@th\sbox\tw@{$#1($}%
  \hss\resizebox{#2}{\wd\tw@}{\rotatebox[origin=c]{90}{\upshape(}}\hss$}

\makeatother

    \newcommand{\midsepremove}{\aboverulesep = 0mm \belowrulesep = 0mm}
    \midsepremove
    \newcommand{\midsepdefault}{\aboverulesep = 0.1mm \belowrulesep = 0.1mm}
    \midsepdefault

\usepackage{amsthm}
\theoremstyle{definition}
\newtheorem{definition}{Definition}

\begin{document}
\let\WriteBookmarks\relax
\def\floatpagepagefraction{1}
\def\textpagefraction{.001}

\begin{titlepage}
\begin{center}
        \vspace*{1cm}
            
        \large
        \textbf{ HGCN(O): A Self-Tuning GCN HyperModel Toolkit for \\  Outcome Prediction in Event-Sequence Data}
            
       \vspace{1.5cm}

       \normalsize
       Fang Wang $^{a, *}$, Paolo Ceravolo $^{a}$, Ernesto Damiani $^{a}$
       \\
       \vspace{6pt}
       \small
       $^a$ \textit{Department of Computer Science, University of Milan, Via Festa del Perdono 7, 20122, Milano, MI, Italy}\\

       \vspace{15pt}

\small
     
       \rule{0.4\textwidth}{0.5pt}\\
       
       \textit{$^*$Corresponding Author}\\
       
\textit{Email addresses: florencewong@protonmail.com (Fang Wang),} \\
\textit{ paolo.ceravolo@unimi.it (Paolo Ceravolo),}\\
\textit{ ernesto.damiani@unimi.it (Ernesto Damiani)}

       \vfill
        
\end{center}

\end{titlepage}

\shorttitle{HGCN(O)}    

\shortauthors{Wang, Ceravolo and Damiani}  

\title [mode = title]{HGCN(O): A Self-Tuning GCN HyperModel Toolkit for Outcome Prediction in Event-Sequence Data}  

\author[1]{Fang Wang}[orcid=0000-0002-7349-8888]

\cormark[1]

\ead{florencewong@protonmail.com}

\credit{Conceptualization, Methodology, Software, Validation, Formal analysis, Investigation, Data curation, Writing - original draft, Writing - review and editing, Visualization}

\affiliation[1]{organization={Department of Computer Science, University of Milan},
            addressline={Via Festa del Perdono 7}, 
            city={Milano},
            citysep={}, %
            postcode={20122}, 
            state={MI},
            country={Italy}}
            
\author[1]{Paolo Ceravolo}[orcid=0000-0002-4519-0173]
\ead{paolo.ceravolo@unimi.it}
\credit{Conceptualization, Data Curation, Validation, Writing - review and editing}

\author[1]{Ernesto Damiani}[orcid=0000-0002-9557-6496]

% Email id of the second author
\ead{ernesto.damiani@unimi.it}

\credit{Supervision, Conceptualization, Methodology, Writing - review and editing, Project administration}

\cortext[1]{Corresponding author}

\begin{abstract}
\textsc{Outcome prediction from event sequence data requires effective use of information at two distinct semantic levels: attributes of individual events and attributes of the complete sequence. Although graph models can explicitly represent temporal relations between events, existing approaches commonly rely on predefined representations and manually configured architectures, with limited consideration of how event and sequence information should be incorporated at their corresponding GNN levels or how GNN models should be fitted to different datasets without manual tuning. We propose HGCN(O), an end to end self tuning graph toolkit for hierarchical information representation and automated model configuration. Its core T-GCN architecture processes event information at the node level and sequence information at the graph level, while O-GCN provides a one level reference for early integration of sequence attributes. TP-GCN and TE-GCN extend T-GCN with a pseudo informative duration embedding and a learned activity embedding, respectively. HGCN(O) integrates these architecture specific designs with hypermodel optimization to jointly determine model structure and training configuration. The core T-GCN achieves F1 scores of 1.0000 on two balanced benchmarks and a weighted F1 of 0.8639 on a highly imbalanced dataset, compared with 0.8318 for GAT and 0.8453 for TransformerConv. Prefix and ablation analyses further demonstrate its effectiveness for outcome prediction from incomplete sequences and quantify the contributions of temporal edge weights and graph level sequence attributes.}
\end{abstract}

%\includegraphics[width=\textwidth]{graph abstract.pdf}
%\item HGCN(O) self tunes GNN architectures for event sequence outcome prediction
%\item Two level T-GCN preserves event and sequence information at corresponding GNN levels
%\item Pseudo informative duration embedding captures activity duration patterns
%\item Auto-configured T-GCN outperforms fixed GAT and TransformerConv baselines
%\item Prefix and ablation analyses support early prediction and structural design


\maketitle

\section{Introduction}
\label{sec:Intro}

Outcome prediction from event sequence data aims to infer a predefined outcome from an observed sequence of events. A major application is Predictive Process Monitoring (PPM) \cite{ceravolo2024predictive}, where the objective is to predict the outcome of a business process from a partially observed event sequence \cite{maggi2014predictive,pika2016evaluating,teinemaa2019outcome}. Outcome prediction is commonly formulated as a classification problem in which each event sequence is assigned to one of several predefined outcomes according to the observed events. Traditional machine learning methods, including Random Forest, XGBoost, Decision Trees, and support vector machines, provided effective early solutions to this task \cite{leontjeva2015complex,senderovich2017intra,grigori2001improving,grigori2004business,castellanos2005predictive,kang2012periodic}. As event sequence behavior became more complex, deep learning shifted the focus from manually defined features toward representations learned directly from event sequences. Recurrent neural networks and LSTM models improved the capture of sequential dependencies \cite{hinkka2019classifying,kratsch2021machine}, while attention based recurrent models further refined the representation of relevant temporal information \cite{wang2019outcome}. Convolutional approaches extended this progression by transforming event sequences into image based representations and learning discriminative latent features for outcome prediction \cite{pasquadibisceglie2020orange}.

Despite these advances, most sequence based approaches model event sequences primarily as ordered observations or transformed feature representations, capturing temporal dependencies while leaving structural relations among events implicit. Graph based learning provides an explicit means of representing these relations \cite{bellandi2022graph}. In PPM, nodes can encode event level attributes, graph edges can represent temporal relations between consecutive events, and graph pooling provides a natural transition from event level representations to information describing the complete sequence. This structure is particularly suitable for event sequence data, where event level and sequence level attributes coexist but represent different semantic levels. Existing graph based process prediction studies, however, have largely focused on particular graph representations or predefined GNN architectures, while the joint organization of temporal relations, event attributes, and sequence attributes remains less systematically addressed.

Recent industrial process monitoring has also moved toward integrated learning frameworks that combine statistical, neural, causal, and fusion based methods for fault detection and diagnosis \cite{ali2025hybrid,ali2026hybrid,ali2026diglpp,ali2025novel,ali2024advance,ali2025intelligent}. This broader direction highlights the value of combining complementary information and learning mechanisms within a unified predictive system. For graph based outcome prediction from event sequence data, such integration introduces a second challenge. Model performance depends not only on the graph representation but also on architectural and training choices, including the graph convolution operator, network depth and width, aggregation, regularization, optimizer, and learning rate strategy \cite{khemani2024review}. These choices are commonly fixed manually, although their effectiveness can vary across datasets, input representations, and class distributions. This creates a need for a framework that addresses graph representation and model configuration jointly rather than treating them as separate design problems.

To address these limitations, we propose HGCN(O), an end to end self tuning graph toolkit for outcome prediction in event sequence data. HGCN(O) builds on established graph learning and hyperparameter optimization techniques while jointly addressing graph construction, information representation, architecture design, and automated model configuration. The main contributions are as follows:

\begin{itemize}[leftmargin=*,noitemsep,nolistsep]

\item \textbf{End to end graph prediction toolkit:} HGCN(O) covers the complete prediction pipeline, from transforming event sequences into attributed graphs to learning graph representations, integrating sequence level information, configuring alternative graph architectures, and producing outcome predictions. These stages are handled within a single toolkit rather than through separately designed graph construction, modeling, and optimization procedures.

\item \textbf{Hierarchical graph architecture:} HGCN(O) departs from flat feature integration by separating event level relational learning from sequence level information integration. Event level information is learned through graph convolution over temporally weighted event relations, while sequence level attributes are introduced after graph pooling. This preserves the semantic distinction between event and sequence information while allowing local relational structure and global graph level information to contribute to the same prediction model.

\item \textbf{Unified representation framework:} HGCN(O) organizes alternative event representations within a common graph architecture rather than treating them as independent model designs. Beyond the core T-GCN model and the one level O-GCN reference, TP-GCN and TE-GCN provide two complementary mechanisms for incorporating additional event information: a pseudo informative duration embedding that captures the relative importance of activity and duration patterns, and a learned activity embedding. This formulation allows different forms of raw, derived, and learned event information to be incorporated within the same framework while preserving the underlying two level graph structure.

\item \textbf{Adaptive hypermodel configuration:} HGCN(O) replaces fixed architectural and training settings with architecture specific automated configuration. The hypermodel jointly adapts graph depth and width, aggregation, dense structure, regularization, optimization, and training strategy according to the instantiated architecture, graph convolution operator, input representation, and dataset. This allows a common model family to adapt to different event sequence characteristics and class distributions without manually designing a separate GNN configuration for each dataset.

\item \textbf{Framework level predictive capability:} HGCN(O) is evaluated as an integrated framework rather than through final accuracy alone. Comparisons with attention based GNNs, balanced and imbalanced datasets, incomplete sequence prefixes, and structural ablations assess its predictive strength, adaptive configuration, capability under incomplete information, and dependence on the proposed graph design beyond automated hyperparameter optimization.
\end{itemize}

As an end to end toolkit, the original contributions of HGCN(O) are the two level graph design, the pseudo informative duration embedding introduced in TP-GCN, their organization within the HGCN(O) model family, and the architecture specific hypermodel configuration. Within the framework, GCNConv, GraphConv, graph pooling, learned activity embeddings, and Optuna are adopted as established building blocks to implement the proposed hierarchical representation and adaptive prediction pipeline. The methodological novelty therefore resides in the framework architecture and integration logic rather than in any individual graph convolution or optimization component.

This paper is classified as a methodological framework study for graph based outcome prediction from event sequence data, with empirical validation in predictive process monitoring. The remainder of the paper is organized as follows. Section~\ref{sec:RL} reviews related work, Sections~\ref{sec:pre} and~\ref{sec:method} present the graph representation and HGCN(O) methodology, Section~\ref{sec:data} describes the datasets and preprocessing, Section~\ref{sec:Exp} describes the experimental evaluation, Section~\ref{sec:RE} reports and discusses the results, and Section~\ref{sec:conclu} concludes the paper.

\section{Related Work}
\label{sec:RL}
This section reviews three lines of research relevant to HGCN(O): graph representation learning and graph neural networks, graph based predictive process monitoring, and AutoML for graph learning. These areas provide the methodological, application, and automated configuration context for the proposed framework.

\subsection{Graph Representation Learning and Graph Neural Networks}
\label{subsec:RL-gnn}
Graph representation learning provides a general framework for modeling entities and their relations. Earlier approaches relied on graph kernels, matrix factorization, and shallow embedding methods such as DeepWalk, Node2Vec, and LINE \cite{nikolentzos2021graph,perozzi2014deepwalk,grover2016node2vec,tang2015line}. Graph neural networks (GNNs) advanced this direction by integrating graph structure and node attributes through message passing. Among the earliest deep architectures were recurrent GNNs, which iteratively updated node states using shared transformations and established the basis for direct processing of graph structured data \cite{scarselli2008graph,zhang2019shne,wang2017topological}. Graph autoencoders later introduced encoder and decoder architectures to derive latent representations that preserve both structural and feature information \cite{wang2016structural,tu2018structural}.

Modern GNN research has developed several message passing operators with different aggregation and interaction mechanisms. GCN and GraphConv perform local neighborhood aggregation over sparse graph structures and support weighted relations, making them suitable for event sequence graphs in which temporal edge weights are represented on observed edges \cite{kipf2016semi,morris2019weisfeiler}. GIN strengthens the discriminative capacity of neighborhood aggregation and reaches the expressive power of the Weisfeiler Lehman test within its message passing class \cite{xu2018powerful}, while GINE extends this formulation to incorporate edge attributes \cite{hu2019strategies}. Attention based operators instead learn neighbor specific interaction weights. GAT applies attention over connected nodes, whereas TransformerConv uses transformer style attention within local graph neighborhoods and can incorporate edge information into the attention computation \cite{velivckovic2017graph,shi2020masked}. Despite their different formulations, these operators remain primarily based on local message passing over observed graph connectivity and represent the principal GNN families considered in this work.

A distinct research direction extends attention beyond local message passing through Graph Transformers. Because transformer attention does not inherently encode graph topology, structural and positional information is commonly introduced explicitly \cite{yao2020heterogeneous,nguyen2022universal,peng2022rethinking,hussain2022global,khemani2024review}. Graphormer incorporates node centrality, shortest path distance, and edge information directly into the attention mechanism \cite{ying2021transformers}. GraphGPS combines local message passing with global attention within a unified architecture \cite{rampavsek2022recipe}, while Exphormer introduces sparse attention structures to reduce the computational demands of global graph attention \cite{shirzad2023exphormer}. Related studies have further investigated structural encoding and efficient attention mechanisms for graph representation learning \cite{canturk2023graph}. More recently, GGATN \cite{wang2026graph} combines a global graph structure encoder with cross attention between graph and sequence representations for full sequence generation. These developments extend graph modeling from neighborhood based propagation toward broader structural interaction, while also introducing additional architectural and training choices that must be configured for a particular dataset and task.

\subsection{Graph Based Predictive Process Monitoring}
Graph based learning has increasingly been applied to predictive process monitoring as an alternative to representing event sequences solely through fixed sequential encodings. Early GCN based studies established the feasibility of modeling process instances as graphs for prediction tasks \cite{chiorrini2021exploiting}. Subsequent work expanded graph representations and architectures to capture richer process information, including attention over sequential graphs \cite{deng2024enhancing}, heterogeneous graph structures and explainable GNNs \cite{pasquadibisceglie2024sf,harl2020explainable}, and graph based formulations for related process prediction tasks \cite{Bellandi2024247}. Other studies have focused on integrating graph structure with temporal sequence modeling. TACO \cite{rama2023embedding} combines graph convolution with recurrent neural networks for next event prediction, while temporal modulation and transition semantics have been encoded directly on graph edges for the same task \cite{wang2025time}. Collectively, these studies show that explicit structural and relational modeling can enrich process prediction beyond conventional sequence representations. However, recent reviews continue to identify representation design, heterogeneous information integration, early prediction, and model adaptability as open challenges in predictive process monitoring \cite{ceravolo2024predictive}.

Related developments in industrial process monitoring further illustrate the move toward integrated learning frameworks. The panoramic review by Ali et al. \cite{ali2025hybrid} organizes advanced process monitoring into data driven, model driven, and fusion driven paradigms and identifies graph based methods, Bayesian optimization, hybrid deep architectures, scalability, and interpretability as important research directions. Recent studies reflect this broader trend by combining complementary statistical, neural, causal, and fusion mechanisms for fault detection and diagnosis. These include integrated machine learning for industrial process monitoring \cite{ali2024advance}, dynamic explainable fusion for fault detection and root cause analysis \cite{ali2025novel}, multi model fusion of statistical and neural representations \cite{ali2025intelligent}, and reconstruction based Bayesian fusion for dynamic monitoring \cite{ali2025process}. More recent approaches further combine causal inference with neuro fuzzy learning and graph based causal analysis \cite{ali2026hybrid} or integrate structure preserving projection with Kolmogorov Arnold Networks for robust fault detection \cite{ali2026diglpp}. Although these methods primarily address industrial sensor processes rather than business event sequences, they demonstrate the value of integrating heterogeneous information and complementary learning mechanisms within a unified monitoring framework.

\subsection{AutoML for Graph Learning}
AutoML has developed as a general approach for reducing manual model configuration through automated hyperparameter optimization, model selection, and neural architecture search. Recent reviews show a continuing shift from tuning isolated parameters toward jointly adapting architectural and training choices to the characteristics of the data and prediction task \cite{barbudo2023eight,heuillet2024efficient,wang2025auto}. Practical systems such as FLAML further emphasize computational efficiency by performing automated search under limited resource budgets \cite{wang2022automated}. These developments establish AutoML as a general framework for adapting predictive models without relying on manually fixed configurations.

This need for automated configuration is particularly pronounced in graph learning, where model performance depends on interacting choices involving graph convolution operators, network depth and width, aggregation, regularization, optimization, and training strategy. Graph AutoML addresses this complexity through automated GNN design and configuration. Early graph neural architecture search explored alternative message passing architectures systematically \cite{oloulade2021graph}, while later methods focused on scalability and search efficiency. PASCA considers architecture search under a scalable graph learning paradigm \cite{zhang2022pasca}, and AutoGT extends automated design to Graph Transformer architectures \cite{zhang2023autogt}. More recent approaches reduce architecture evaluation cost through training free or lightweight search \cite{xu2023not}, while large language models have also been introduced to guide graph architecture search \cite{wang2025graph}. Despite this progress, most existing methods target general graph architecture search, whereas graph based predictive process monitoring still largely relies on predefined GNN structures and manually specified training configurations.

\section{Preliminaries and Event Sequence Graph Representation}
\label{sec:pre}

We represent each event sequence as a directed attributed graph. Events form the graph nodes, consecutive events are connected by directed edges, and temporal information is encoded through edge weights. Attributes describing individual events are represented at the node level, while attributes describing the complete sequence are retained as graph level features.

Let $G_j=(X_1,\ldots,X_{n_j})$ denote an event sequence containing $n_j$ events, where $X_i$ is the $i$th event in the observed sequence.

\begin{definition}[\textbf{Node Attribute Vector}]
\label{def:node_attributes}
For an event $X_i$, let $\mathcal{A}_i$ denote its activity label, $U_i=(U_{ik})_{k=1}^{d_U}$ the universal attributes available across events, and $B_i=(B_{ik})_{k=1}^{d_B}$ the activity specific attributes. The unencoded node attribute vector is $\widetilde{\mathbf{v}}_{N_i}=[\mathcal{A}_i,U_i,B_i]$. After encoding, it is represented by $\mathbf{v}_{N_i}\in\mathbb{R}^{d_N}$. When available, the activity duration $T_i^d$ is included in $U_i$.
\end{definition}

Activity specific attributes are applicable only to particular activities. For example, \textit{``credit score''} may be relevant to a \textit{``check insurance''} activity but not to a \textit{``register''} activity.

\begin{definition}[\textbf{Activity Label Vector}]
\label{def:activity_attribute}
The ordered activity labels of an event sequence $G_j$ are represented by the vector $\mathbf{a}_{G_j}=(\mathcal{A}_i)_{i=1}^{n_j}$, where $n_j$ denotes the number of events in $G_j$. The activity label identifies the event type and may correspond to an activity name, code, or other categorical identifier provided by the dataset.
\end{definition}

\begin{definition}[\textbf{Activity Duration Vector}]
\label{def:activity_duration}
For an event $X_i$ with start timestamp $T_i^s$ and completion timestamp $T_i^c$, its duration is defined as $T_i^d=T_i^c-T_i^s$. When duration information is available, the activity durations of the event sequence $G_j$ are represented by the vector $\mathbf{t}_{G_j}^{d}=(T_i^d)_{i=1}^{n_j}$.
\end{definition}

\begin{definition}[\textbf{Graph Attribute Vector}]
\label{def:graph_attribute_vector}
Let $F_j=(F_{jk})_{k=1}^{d_F}$ denote the attributes associated with the complete event sequence $G_j$. After encoding, they form the graph attribute vector $\mathbf{v}_{G_j}\in\mathbb{R}^{d_G}$, where $d_G$ denotes the dimension of the encoded graph level feature space.
\end{definition}

Categorical node and graph attributes are encoded using one-hot encoding, while numerical attributes are normalized using min-max scaling. For activity specific node attributes that are not applicable to a given event, numerical values are replaced by the corresponding attribute median, whereas categorical values are assigned $-1$. The value $-1$ is also used for padded entries, which are masked during training.

\begin{definition}[\textbf{Temporal Edge Weight}]
\label{def:edge_weight}
For consecutive events $X_i$ and $X_{i+1}$, let $\widetilde{w}_{i\rightarrow i+1}=T_{i+1}^{s}-T_i^{s}$ denote the temporal gap between their start times. Thus, $\widetilde{w}_{i\rightarrow i+1}=0$ when the two events share the same start timestamp. The temporal gaps are normalized to $[0,1]$ using min-max scaling, yielding the temporal edge weights $w_{i\rightarrow i+1}$. The temporal edge weights of event sequence $G_j$ are represented by $\mathbf{w}_{G_j}=(w_{i\rightarrow i+1})_{i=1}^{n_j-1}$.
\end{definition}

\begin{definition}[\textbf{Directed Edge Index}]
\label{def:edge_index}
The directed edge set of $G_j$ is $\mathcal{E}_{G_j}=\{(i,i+1)\mid 1\leq i<n_j\}$, connecting the node representing each event to the node representing its immediate successor. The corresponding edge index is $\mathbb{E}_{G_j}\in\mathbb{N}^{2\times(n_j-1)}$, where each column contains the source and destination node indices of one directed edge.
\end{definition}

\begin{definition}[\textbf{Event Sequence Graph Representation}]
\label{def:graph_representation}
The encoded node attribute vectors of $G_j$ form the node feature matrix
\[
\mathbb{V}_{G_j}
=
(\mathbf{v}_{N_i})_{i=1}^{n_j}
\in
\mathbb{R}^{n_j\times d_N},
\]
where $d_N$ is the dimension of an encoded node vector. The graph representation of the event sequence is defined as
\[
\mathcal{G}_j
=
\left(
\mathbb{V}_{G_j},
\mathbb{E}_{G_j},
\mathbf{w}_{G_j},
\mathbf{v}_{G_j}
\right),
\]
where $\mathcal{G}_j$ comprises the node feature matrix, directed edge structure, temporal edge weights, and graph level attribute vector.
\end{definition}

Some studies \cite{teinemaa2019outcome} determine the direction of edges between events based on their completion times. In our representation, directed edges preserve the observed event order, while temporal edge weights are computed from the differences between event start times. Since an activity may continue after the subsequent activity has started, its duration $T_i^d$ is retained separately as a universal node attribute. This distinction allows the graph representation to capture both the temporal spacing between consecutive events and the duration of individual activities.

\section{Methodology}
\label{sec:method}

\subsection{Graph Convolution Operators}
\label{subsec:GCN}

HGCN(O) employs two graph convolution operators, \textbf{GCNConv} and \textbf{GraphConv}, to propagate information over the event sequence graph. Both operators process the node representations together with the directed edge structure and temporal edge weights defined in Section~\ref{sec:pre}, but differ in how information from neighboring nodes is aggregated.

\textbf{GCNConv} follows the normalized graph convolution introduced by \citet{kipf2016semi}. Let $\mathbb{V}_{G_j}^{(l)} \in \mathbb{R}^{n_j \times d_l}$ denote the node representation matrix at layer $l$, with $\mathbb{V}_{G_j}^{(0)}=\mathbb{V}_{G_j}$. The layer update is given by
\begin{equation}
\mathbb{V}_{G_j}^{(l+1)}
=
\sigma
\left(
\widetilde{D}^{-\frac{1}{2}}
\widetilde{A}_{w}
\widetilde{D}^{-\frac{1}{2}}
\mathbb{V}_{G_j}^{(l)}
W^{(l)}
\right),
\label{eq:gcnconv}
\end{equation}
where $W^{(l)}$ is a learnable weight matrix and $\sigma$ denotes the activation function. The matrix $A_w$ is the weighted adjacency matrix induced by the directed edges and their temporal edge weights $w_{i\rightarrow i+1}$. Self loops are introduced through $\widetilde{A}_w=A_w+I$, and $\widetilde{D}$ denotes the corresponding degree matrix. The degree normalization in Eq.~\eqref{eq:gcnconv} scales the contribution of neighboring nodes according to the weighted graph structure, providing normalized message passing across successive graph convolution layers.

\textbf{GraphConv}~\cite{morris2019weisfeiler} uses a different propagation mechanism by applying separate transformations to the representation of the central node and to the information aggregated from its neighbors. For the node corresponding to event $X_i$, the update at layer $l$ is expressed as
\begin{equation}
\mathbf{v}_{i}^{(l+1)}
=
\sigma
\left(
W_{r}^{(l)}\mathbf{v}_{i}^{(l)}
+
W_{n}^{(l)}
\operatorname{AGGR}
\left\{
w_{k\rightarrow i}\mathbf{v}_{k}^{(l)}
:
X_k \in \mathcal{N}(X_i)
\right\}
\right),
\label{eq:graphconv}
\end{equation}
where $\mathbf{v}_{i}^{(l)}$ denotes the representation of the node corresponding to event $X_i$ at layer $l$, $\mathcal{N}(X_i)$ denotes its neighborhood, and $W_r^{(l)}$ and $W_n^{(l)}$ are learnable weight matrices for the central node and neighboring representations, respectively. The temporal edge weight $w_{k\rightarrow i}$ scales the contribution of each neighboring node before aggregation.

In HGCN(O), the GraphConv aggregation operator is treated as a tunable architectural parameter, with addition, mean, and maximum aggregation considered independently at each graph convolution layer. GCNConv and GraphConv therefore provide two complementary propagation mechanisms under the same event sequence representation: GCNConv performs degree normalized propagation over the weighted graph structure, whereas GraphConv separates the central node transformation from weighted neighborhood aggregation and allows the aggregation operation itself to be optimized. Both operators are retained throughout the HGCN(O) architecture family and are tuned independently according to the characteristics of each dataset.

\subsection{HGCN(O) Architecture Family}
\label{subsec:HGCN}

HGCN(O) is organized around the two level T-GCN architecture, which preserves the distinction between node level and graph level information. O-GCN provides a one level reference configuration for early integration of graph attributes, while TP-GCN and TE-GCN extend T-GCN with additional representations derived from activity duration and activity labels, respectively. Each architecture can be instantiated using either GCNConv or GraphConv, resulting in eight model configurations.

TP-GCN introduces a pseudo informative duration embedding to represent activity duration as a local temporal property distinct from the temporal relation between consecutive events captured by the edge weights. The additional branch also provides a mechanism for incorporating auxiliary structured representations derived from event information without altering the core two level architecture. TE-GCN instead introduces a learned activity embedding branch that maps activity labels into a dedicated latent representation. Depending on the available event information, this branch is incorporated into the core T-GCN structure directly or alongside the pseudo embedding branch, while TE-GCN remains the activity embedding extension within the HGCN(O) model family. The following subsections describe the four architectures in this order.

\subsubsection{Two Level Core GCN (T-GCN)}
\label{sbsbsec:tgcn}
The \textbf{Two Level GCN (T-GCN)} constitutes the core architecture of HGCN(O). It preserves node level and graph level information as separate inputs during representation learning. For an event sequence $G_j$, the node feature matrix $\mathbb{V}_{G_j}$, directed edge structure $\mathbb{E}_{G_j}$, and temporal edge weights $\mathbf{w}_{G_j}$ are processed through $L_G$ graph convolution layers, instantiated using either GCNConv or GraphConv, producing the node representation matrix $\mathbb{V}_{G_j}^{(L_G)}$. A graph pooling operation then aggregates the learned node representations into a graph embedding $\mathbf{z}_{G_j}=\operatorname{POOL}(\mathbb{V}_{G_j}^{(L_G)})$.

In parallel, the encoded graph attribute vector $\mathbf{v}_{G_j}$ is processed through $d$ fully connected layers, yielding the transformed graph level representation $\mathbf{v}_{G_j}^{(d)}$. The two representations are subsequently combined as $ \mathcal{Z}_{G_j} = \left[ \mathbf{z}_{G_j}, \mathbf{v}_{G_j}^{(d)} \right]$.

The fused representation $\mathcal{Z}_{G_j}$ is processed through a sequence of fully connected layers and a final classification layer. Let $\mathcal{Y}$ denote the set of outcome classes. The classifier produces $f_{\theta}(\mathcal{Z}_{G_j})\in\Delta^{|\mathcal{Y}|-1}$, where $\theta$ denotes the learnable model parameters and $\Delta^{|\mathcal{Y}|-1}$ is the probability simplex over the outcome classes. The predicted outcome is
\[
\hat{y}_{G_j}
=
\arg\max_{c\in\mathcal{Y}}
f_{\theta}(\mathcal{Z}_{G_j})_c.
\]

\subsubsection{One Level Reference GCN (O-GCN)}
\label{sbsbsec:OGCN}

The \textbf{One Level GCN (O-GCN)} provides the reference configuration for integrating node and graph information before graph convolution. For an event sequence $G_j$ containing $n_j$ events, the graph attribute vector $\mathbf{v}_{G_j} \in \mathbb{R}^{d_G}$ is replicated across all nodes and concatenated with the node feature matrix $\mathbb{V}_{G_j} \in \mathbb{R}^{n_j \times d_N}$:
\[
\mathbb{F}_{G_j}
=
\mathbf{1}_{n_j}\mathbf{v}_{G_j}^{\intercal}
\in
\mathbb{R}^{n_j \times d_G},
\qquad
\mathbb{V}_{G_j}^{O}
=
\left[
\mathbb{V}_{G_j},
\mathbb{F}_{G_j}
\right]
\in
\mathbb{R}^{n_j \times (d_N+d_G)},
\]
where $\mathbf{1}_{n_j}$ is a vector of ones. The resulting augmented event graph is $\mathcal{G}_{j}^{O}=(\mathbb{V}_{G_j}^{O},\mathbb{E}_{G_j},\mathbf{w}_{G_j})$.

The graph $\mathcal{G}_{j}^{O}$ is processed through $L_G$ graph convolution layers instantiated using either GCNConv or GraphConv, yielding the final node representation matrix $\mathbb{V}_{G_j}^{O(L_G)}$. A graph pooling operation produces $\mathbf{z}_{G_j}^{O}=\operatorname{POOL}(\mathbb{V}_{G_j}^{O(L_G)})$. The pooled representation is subsequently processed through fully connected layers and the final classifier to obtain the predicted outcome $\hat{y}_{G_j}$.

Compared with T-GCN, O-GCN integrates graph attributes at the node level before message passing, providing a reference for assessing the effect of preserving node level and graph level information separately.

\subsubsection{Two Level Pseudo Embedding GCN (TP-GCN)}
\label{sbsbsec:TPGCN}

The \textbf{Two Level Pseudo Embedding GCN (TP-GCN)} extends T-GCN by introducing an auxiliary pseudo embedding branch derived from activity duration. Although the raw duration $T_i^d$ is already included as a universal node attribute, the pseudo embedding provides a different representation of the same temporal information. Instead of treating duration as an independent numerical value, it captures the relative importance of activity duration patterns across event sequences. The duration based pseudo representation introduced in HGCN(O) was subsequently extended to multilevel LSTM hypermodels \cite{wang2024lstm,wang2025leveraging}.

The pseudo informative duration embedding is constructed from the joint occurrence of activity labels and activity durations. Each duration $T_i^d$ is first assigned to a duration bin, and the resulting pair $(\mathcal{A}_i,b_i)$ is treated as a discrete activity duration pattern. These patterns are then weighted using Term Frequency Inverse Document Frequency (TF-IDF), where each event sequence is treated as a document and each activity duration pattern as a term. The resulting weights form the pseudo embedding representation used by TP-GCN. Algorithm~\ref{alg:pseudo_embedding} summarizes the construction procedure.

\begin{algorithm}[htbp]
\caption{Pseudo Informative Duration Embedding Matrix Construction}
\label{alg:pseudo_embedding}
\begin{algorithmic}[1]

\REQUIRE Event sequences $\{G_j\}$ with activity labels $\mathcal{A}_i$ and activity durations $T_i^d$
\ENSURE Node aligned pseudo informative duration embedding matrix $\mathbb{P}_{G_j}$ for each event sequence $G_j$

\STATE Generate candidate combinations of the duration cutoff $T_{\mathrm{cut}}$ and the number of quantile bins $N_q$

\FOR{each candidate $(T_{\mathrm{cut}},N_q)$}
    \STATE Partition activity durations into two ranges using $T_{\mathrm{cut}}$

    \FOR{each event $X_i$}
        \IF{$T_i^d < T_{\mathrm{cut}}$}
            \STATE Assign $T_i^d$ to an individual duration bin
        \ELSE
            \STATE Assign $T_i^d$ to one of $N_q$ quantile bins
        \ENDIF
    \ENDFOR

    \STATE Compute the frequencies of the resulting quantile bins
    \STATE Compute the balance score as the difference between the maximum and minimum quantile bin frequencies
\ENDFOR

\STATE Select the $(T_{\mathrm{cut}},N_q)$ combination with the minimum balance score
\STATE Assign each event $X_i$ to its resulting duration bin $b_i$

\STATE Form the set of activity duration patterns $\{(\mathcal{A}_i,b_i)\}$
\STATE Treat each distinct $(\mathcal{A}_i,b_i)$ pair as a term

\FOR{each event sequence $G_j$}
    \STATE Treat $G_j$ as a document
    \STATE Compute the TF-IDF weight of each activity duration term for $G_j$
    \STATE Construct the node aligned pseudo embedding matrix
    $\mathbb{P}_{G_j} = (\mathbf{p}_i)_{i=1}^{n_j}$
\ENDFOR

\RETURN $\{\mathbb{P}_{G_j}\}$

\end{algorithmic}
\end{algorithm}

For an event sequence $G_j$ with $n_j$ events, the resulting pseudo embedding vectors form the matrix
\[
\mathbb{P}_{G_j}
=
(\mathbf{p}_i)_{i=1}^{n_j}
\in
\mathbb{R}^{n_j \times d_P},
\]
where $\mathbf{p}_i$ denotes the pseudo embedding vector associated with the activity duration pattern of event $X_i$, and $d_P$ denotes the dimension of the pseudo embedding representation. Using the same directed edge structure and temporal edge weights as the original event sequence graph, the auxiliary graph is defined as $\mathcal{G}_{j}^{P}=(\mathbb{P}_{G_j},\mathbb{E}_{G_j},\mathbf{w}_{G_j})$.

The original node feature matrix $\mathbb{V}_{G_j}$ and the pseudo embedding matrix $\mathbb{P}_{G_j}$ are processed independently through graph convolution branches, producing the node representations $\mathbb{H}_{G_j}^{N}$ and $\mathbb{H}_{G_j}^{P}$, respectively. These representations are combined at the node level as $ \mathbb{H}_{G_j}^{TP} = \left[
\mathbb{H}_{G_j}^{N},
\mathbb{H}_{G_j}^{P}
\right]$. 
The combined representation is further processed through graph convolution layers and pooled to obtain $\mathbf{z}_{G_j}^{TP}$. From this stage, TP-GCN follows the core T-GCN architecture. The graph attribute vector $\mathbf{v}_{G_j}$ is processed through its fully connected branch and combined with $\mathbf{z}_{G_j}^{TP}$ as
$ \mathcal{Z}_{G_j}^{TP}
=
\left[
\mathbf{z}_{G_j}^{TP},
\mathbf{v}_{G_j}^{(d)}
\right]$. The fused representation is then processed through the final classification layers to produce $\hat{y}_{G_j}$.

\subsubsection{Two Level Embedding GCN (TE-GCN)}
\label{sbsbsec:TEGCN}

The \textbf{Two Level Embedding GCN (TE-GCN)} introduces a learned representation of the activity labels $\mathcal{A}_i$. Since the activity label identifies the semantic type of an event, it plays a central role in characterizing an event sequence. Rather than representing this information only through its encoded form in the node attribute vector, TE-GCN maps each activity label to a trainable embedding space.

For an event sequence $G_j$, the activity labels are transformed through an embedding layer to form
\[
\mathbb{V}_{G_j}^{E}
=
(\mathbf{e}_i)_{i=1}^{n_j}
\in
\mathbb{R}^{n_j \times d_E},
\]
where $\mathbf{e}_i$ denotes the learned embedding of activity $\mathcal{A}_i$ and $d_E$ is the embedding dimension. The activity embedding matrix $\mathbb{V}_{G_j}^{E}$ is associated with the original directed edge structure $\mathbb{E}_{G_j}$ and temporal edge weights $\mathbf{w}_{G_j}$ to form the auxiliary graph $\mathcal{G}_{j}^{E}=(\mathbb{V}_{G_j}^{E},\mathbb{E}_{G_j},\mathbf{w}_{G_j})$.

In this study, the activity embedding is examined in two configurations. First, it is introduced directly into the core T-GCN architecture as an additional event representation. Separate graph convolution branches are then applied to the original node feature matrix $\mathbb{V}_{G_j}$ and the activity embedding matrix $\mathbb{V}_{G_j}^{E}$, yielding the node representations $\mathbb{H}_{G_j}^{N}$ and $\mathbb{H}_{G_j}^{E}$, respectively. These representations are combined as $ \mathbb{H}_{G_j}^{TE}
=
\left[
\mathbb{H}_{G_j}^{N},
\mathbb{H}_{G_j}^{E}
\right] $. The combined representation is then processed through subsequent graph convolution layers and graph pooling. Second, when the pseudo informative duration embedding is available, the activity embedding branch is added to the TP-GCN structure. In this configuration, the TE-GCN node representation is
$ \mathbb{H}_{G_j}^{TE}
=
\left[
\mathbb{H}_{G_j}^{N},
\mathbb{H}_{G_j}^{P},
\mathbb{H}_{G_j}^{E}
\right]$. The original node representation, pseudo informative duration embedding, and learned activity embedding are therefore incorporated before subsequent graph convolution and pooling.

For both configurations, TE-GCN retains the graph level structure of the core T-GCN model. Graph pooling produces a graph embedding, while the graph attribute vector $\mathbf{v}_{G_j}$ is processed independently and fused with the pooled representation for outcome classification.

Table~\ref{tab:model_variants} summarizes the structural differences among the four HGCN(O) model variants, including their information inputs, sequence attribute integration, auxiliary representations, and defining formulations.

\begin{table}[!htbp]
\centering
\caption{Comparative Summary of HGCN(O) Model Variants}
\label{tab:model_variants}

\small
\renewcommand{\arraystretch}{1.2}
\setlength{\tabcolsep}{2pt}

\begin{tabular}{
@{}
>{\raggedright\arraybackslash}p{0.07\textwidth}
>{\raggedright\arraybackslash}p{0.08\textwidth}
>{\raggedright\arraybackslash}p{0.17\textwidth}
>{\raggedright\arraybackslash}p{0.13\textwidth}
>{\raggedright\arraybackslash}p{0.13\textwidth}
>{\raggedright\arraybackslash}p{0.18\textwidth}
>{\raggedright\arraybackslash}p{0.18\textwidth}
@{}
}
\toprule

\textbf{Model} &
\textbf{Role} &
\textbf{Event Level Input} &
\textbf{Sequence Level Input} &
\textbf{Sequence Attribute Integration} &
\textbf{Auxiliary Representation} &
\textbf{Key Formulation} \\

\midrule

O-GCN &
Reference &
Event attributes &
Sequence attributes replicated across nodes &
Node level, before graph convolution &
None &
$\mathbb{V}_{G_j}^{O}
=
\left[
\mathbb{V}_{G_j},
\mathbf{1}_{n_j}\mathbf{v}_{G_j}^{\intercal}
\right]$
\\[2pt]

T-GCN &
Core model &
Event attributes &
Sequence attributes &
Graph level, after pooling &
None &
$\mathcal{Z}_{G_j}
=
\left[
\mathbf{z}_{G_j},
\mathbf{v}_{G_j}^{(d)}
\right]$
\\[2pt]

TP-GCN &
Extension &
Event attributes &
Sequence attributes &
Graph level, after pooling &
Pseudo informative duration embedding &
$\mathbb{H}_{G_j}^{TP}
=
\left[
\mathbb{H}_{G_j}^{N},
\mathbb{H}_{G_j}^{P}
\right]$
\\[2pt]

TE-GCN &
Extension &
Event attributes &
Sequence attributes &
Graph level, after pooling &
Learned activity embedding; optional pseudo embedding &
$\mathbb{H}_{G_j}^{TE}
=
\left[
\mathbb{H}_{G_j}^{N},
\mathbb{H}_{G_j}^{E}
\right]$
or
$\left[
\mathbb{H}_{G_j}^{N},
\mathbb{H}_{G_j}^{P},
\mathbb{H}_{G_j}^{E}
\right]$
\\

\bottomrule
\end{tabular}
\end{table}

\subsection{Automated Hypermodel Optimization}
\label{subsec:hyper}

A central component of HGCN(O) is the automated construction and optimization of its model configurations. Rather than assigning a fixed architecture and training configuration to each dataset, HGCN(O) treats every member of the architecture family as a hypermodel whose structure, regularization, and learning strategy are jointly determined from a predefined search space. For a hypermodel $\mathcal{H}$ with configuration $\boldsymbol{\lambda}\in\Lambda$, the optimization process seeks
\[
\boldsymbol{\lambda}^{*}
=
\arg\max_{\boldsymbol{\lambda}\in\Lambda}
\mathcal{J}
\left(
\mathcal{H}_{\boldsymbol{\lambda}}
\right),
\]
where $\Lambda$ denotes the hyperparameter search space and $\mathcal{J}$ denotes the model selection criterion, including the primary validation metric and its associated tie breaking criteria.

The search space spans three complementary aspects of model construction. Architectural parameters determine the number and dimensions of graph convolution and fully connected layers, skip connections, graph pooling, and the neighborhood aggregation operation for GraphConv. For TE-GCN, the activity embedding dimension is also optimized. Representation and regularization parameters include activation functions, batch normalization, dropout, and $\ell_1$ regularization. Training parameters cover the optimizer and its associated parameters, learning rate, weight decay, learning rate scheduler, batch size, and loss function. These parameters are optimized jointly rather than in isolation, allowing the resulting configuration to adapt both the network structure and the learning process to the characteristics of the target dataset. The complete search space is reported in Table~\ref{tab:hyperparameters}.

Optimization is performed using Optuna, which evaluates candidate configurations over successive trials and retains the configuration that best satisfies the model selection objective. The objective is adapted to the class distribution of the dataset. For balanced datasets, validation accuracy is used as the primary criterion. Configurations with equal accuracy are compared using the standard deviation of the validation loss, followed by the validation loss itself. For imbalanced datasets, weighted F1 is used as the primary criterion, with validation loss and its standard deviation used to resolve ties. This criterion allows the optimization process to account for precision and recall across classes while preserving the observed class distribution.

Early stopping is applied within each trial to terminate training when the monitored validation criterion no longer improves, while Optuna pruning terminates unpromising configurations before completing the full training schedule. Automated model configuration, training optimization, and trial management form the self tuning component of HGCN(O), enabling each architecture to be instantiated according to the statistical and structural characteristics of the data rather than through a manually selected parameter configuration.

\begin{table}[!htbp]
\centering
\caption{HGCN(O) Hyperparameter Search Space}
\label{tab:hyperparameters}
\begin{tabularx}{\linewidth}{Xrl}
\toprule
\textbf{Component} & \textbf{Hyperparameter} & \textbf{Search Space} \\
\midrule

\multirow{8}{*}{Architecture}
& Graph convolution layers & 1-5 per applicable branch \\
& Graph convolution hidden units & 32-256 per layer \\
& Fully connected layers & 1-3 per applicable branch \\
& Fully connected hidden units & 32-256 per layer \\
& TE embedding dimension & 10-50 \\
& Graph pooling & mean, max, add \\
& GraphConv aggregation & add, mean, max per layer \\
& Skip connection & on, off \\
\midrule

& Activation & ReLU, LeakyReLU, ELU, Tanh, Softplus, GELU \\
& Batch normalization & on, off \\
Layer & Batch norm momentum & 0.01-0.99 \\
configuration & Batch norm epsilon & $10^{-5}$-$10^{-2}$ \\
& Dropout & on, off; rate: 0.1-0.5 \\
\midrule

\multirow{5}{*}{Training}
& Learning rate & $10^{-8}$-$10^{-1}$ (log scale) \\
& Weight decay & 0-$10^{-2}$ \\
& L1 coefficient & 0-$10^{-3}$ \\
& Batch size & 32, 64, 128, 256, 512 \\
& Loss function & CrossEntropy, MultiMargin \\
\midrule

\multirow{4}{*}{Optimizer}
& Optimizer & Adam, SGD, RMSprop \\
& Adam & $\beta_1$: 0.85-0.99; $\beta_2$: 0.90-0.999 \\
& SGD & momentum: 0-0.9 \\
& RMSprop & $\alpha$: 0-0.999; momentum: 0-0.999; 
epsilon: $10^{-9}$-$10^{-7}$ \\
\midrule

\multirow{7}{*}{LR scheduler}
& StepLR & step size: 1-50; $\gamma$: 0.1-0.9 \\
& ExponentialLR & $\gamma$: 0.85-0.99 \\
& ReduceLROnPlateau & factor: 0.1-0.9; patience: 1-50;
threshold: $10^{-4}$-$10^{-2}$; epsilon: $10^{-8}$-$10^{-4}$ \\
& PolynomialLR & power: 0.1-2.0; total iterations: 2-300 \\
& CosineAnnealingLR & $T_{\max}$: 10-100; $\eta_{\min}$: $10^{-6}$-$10^{-2}$ \\
& CyclicLR & base LR: $10^{-5}$-$10^{-2}$ (log);
max LR: $10^{-3}$-$10^{-1}$ (log); step size up: 5-200 \\
& OneCycleLR & max LR: $10^{-3}$-$10^{-1}$;
$pct\_start$: 0.1-0.5 \\
\bottomrule
\end{tabularx}
\end{table}

Architectural and layer specific hyperparameters are optimized independently for each applicable branch of a given HGCN(O) architecture. Accordingly, only hyperparameters relevant to the instantiated architecture and convolution operator are included in its search space.

\subsection{Baseline GNN Architectures}
\label{subsec:comparative_gnn}

To evaluate HGCN(O) against alternative graph learning mechanisms, we additionally implement a Graph Attention Network (GAT) and a TransformerConv as comparative baselines. These models are not components of the HGCN(O) architecture family and are not treated as hypermodels within its automated optimization framework. To maintain a controlled comparison, both models use the same event sequence representation and available input information as the core T-GCN configuration, while replacing the graph convolution mechanism with their respective attention based graph operators.

The GAT baseline employs attention based neighborhood aggregation, allowing the contribution of neighboring events to be learned during message passing. The TransformerConv baseline uses transformer based graph attention to model interactions among event nodes while preserving the graph structure of the event sequence. The specific configurations used for these comparative models are reported in Section~\ref{subsec:comparative_config}.

\section{Datasets and Preprocessing}
\label{sec:data}

Three datasets are used to evaluate HGCN(O) under different class distributions and event sequence characteristics: BPI12-A, BPI12-O~\cite{BPIC12set}, and Patients. BPI12-A and BPI12-O provide balanced settings with three outcome classes, whereas Patients represents a substantially imbalanced setting with six outcome classes. Table~\ref{tab:datasets} summarizes the main characteristics of the datasets, including event sequence and activity counts, class imbalance, sequence length and duration, and the available node and graph attributes.

BPI12-A and BPI12-O record real world application processes for personal loans and overdrafts. Their outcomes correspond to accepted, declined, and canceled applications, with BPI12-A containing 2,224 event sequences per outcome and BPI12-O containing 802 event sequences per outcome. The Patients dataset represents healthcare related event sequences and exhibits a markedly different class distribution: the majority class accounts for 40.74\% of the event sequences, whereas the minority class accounts for only 1.12\%, corresponding to an imbalance ratio of approximately 36:1. Its richer attribute structure also includes both universal and activity specific node attributes together with multiple graph level attributes.

All datasets contain event start and completion timestamps. Temporal edge weights are computed from consecutive event start times as defined in Section~\ref{sec:pre}. For Patients, activity duration is additionally retained as a node attribute and used to construct the pseudo informative duration embedding employed by TP-GCN. Durations are rounded to the nearest minute; durations below five minutes are assigned individual bins, whereas longer durations are divided into 19 quantile based bins, resulting in approximately 24 bins in total. BPI12-A and BPI12-O contain zero activity durations and therefore do not use the pseudo informative duration embedding.

\begin{table}[!htbp]
\centering
\begin{threeparttable}
\caption{Characteristics of the Evaluation Datasets}
\label{tab:datasets}
\setlength{\tabcolsep}{3pt}

\begin{tabular}{lccccccccc}
\toprule
\textbf{Dataset} & \textbf{\#Seq} & \textbf{\#Act} & \textbf{\#Cls} & \textbf{IR} & \textbf{Med($l$)} & \textbf{Range($l$)} & \textbf{Med($T$)} & \textbf{Graph attr} & \textbf{Node attr} \\
\midrule
BPI12-A & 6,672 & 9 & 3 & $\sim$1:1 & 6 & 3-7 & 14,579 & 1 numerical & universal: 2 categorical \\
BPI12-O & 2,406 & 7 & 3 & $\sim$1:1 & 5 & 4-30 & 20,542 & 1 numerical & universal: 2 categorical \\
\multirow{2}{*}{Patients} & \multirow{2}{*}{2,140} & \multirow{2}{*}{10} & \multirow{2}{*}{6} & \multirow{2}{*}{36:1} & \multirow{2}{*}{7} & \multirow{2}{*}{4-9} & \multirow{2}{*}{8,505} & 3 numerical & specific: 3 numerical; 1 categorical \\
& & & & & & & & 1 categorical & universal: 1 categorical \\
\bottomrule
\end{tabular}

\begin{tablenotes}
\item \#Seq denotes the number of sequences; \#Act denotes the number of distinct activities; \#Cls denotes the number of outcome classes; IR denotes the ratio between the largest and smallest classes; $l$ denotes sequence length; Med($l$) denotes the median sequence length; Range($l$) denotes the minimum and maximum sequence lengths; Med($T$) denotes the median sequence duration in minutes.
\end{tablenotes}

\end{threeparttable}
\end{table}

\section{Experiments}
\label{sec:Exp}
All experiments were conducted on a GPU enabled Dell XPS 15 9570 workstation running Windows 11 Pro, equipped with an Intel Core i9-8950HK processor, an NVIDIA GeForce GTX 1050 Ti with Max-Q Design GPU with 4\,GB of memory, and 32\,GB of RAM. The experimental environment used Python 3.12.12, PyTorch 2.7.1 with CUDA 11.8, and Optuna 4.7.0 \cite{akiba2019optuna} for automated hyperparameter optimization.

\subsection{Experimental Pipeline}
\label{subsec:pipe}

The experimental pipeline is structured in four parts. First, GAT and TransformerConv are used as strong attention based baselines under the same event sequence representation as the core T-GCN model. This comparison examines whether automated configuration of the HGCN(O) framework can achieve competitive or stronger performance using comparatively simple graph convolution operators. Second, the complete HGCN(O) family is studied on the imbalanced Patients dataset through T-GCN, O-GCN, TP-GCN, and TE-GCN with both GCNConv (G) and GraphConv (C), while the six applicable configurations are considered on BPI12-A and BPI12-O. These experiments characterize how the hypermodel adapts architectural and training configurations across input structures, graph operators, and different class distributions. Third, the core T-GCN model is trained and evaluated using 30\%, 50\%, 70\%, and 100\% sequence prefixes to assess outcome prediction under incomplete information on the two balanced benchmark datasets. The analysis focuses on how performance evolves as event sequence information becomes available and how sensitive prediction is to sequence completeness, as reported in Section~\ref{sbsbsec:prefix}. Fourth, an ablation study removes either the temporal edge weights or the graph attribute branch from T-GCN and reoptimizes each resulting model through the same automated hyperparameter search, isolating the contributions of temporal relations and graph level sequence information under independently optimized configurations. Overall, the pipeline is designed to validate HGCN(O) as an integrated framework for outcome prediction across its different components and evaluation settings.

\subsection{HGCN(O) Experimental Setup}
\label{subsec:setup}

HGCN(O) was evaluated using a train/validation/test protocol. Each dataset was divided into 80\% training and 20\% held out test sets using a fixed random seed of 42. Five fold cross validation was conducted within the training partition for hyperparameter optimization. For the Patients dataset, both the initial training/test split and the cross validation folds were stratified by outcome class to preserve the class distribution. For BPI12-A and BPI12-O, a custom length stratified procedure was used to preserve the sequence length distribution and to ensure that all activities occurring in the validation or test sets were represented in the corresponding training data.

Within each fold, the HGCN(O) hypermodel was optimized over 200 Optuna trials, yielding one best configuration per fold. The five retained configurations were subsequently evaluated across the training folds, and the configuration with the best mean validation performance was selected as the final configuration. Each trial was trained for up to 300 epochs with an early stopping patience of 30 epochs and pruning of unpromising trials. The selected configuration was then retrained using the complete 80\% training partition and evaluated on the held out 20\% test set. Final test performance was averaged over three independent runs.

\subsection{Baseline GNN Configuration}
\label{subsec:comparative_config}

GAT and TransformerConv were implemented as fixed comparative baseline models rather than hypermodels. To maintain a controlled comparison with T-GCN, both models preserve the same two level structure: node level information is processed through the graph branch, graph attributes are processed separately through a fully connected branch, and the two representations are combined after graph pooling. The graph convolution operator is replaced by the corresponding GAT or TransformerConv operator, while the remaining architecture is kept fixed. The same temporal edge weights $\mathbf{w}_{G_j}$ used in HGCN(O) are retained in both attention based baselines during message passing.

Both baselines use two graph layers with a hidden dimension of 64, followed by mean graph pooling. The graph attribute branch and the post fusion layer each use 64 hidden units. For the TransformerConv baseline, the first layer uses four attention heads and the second uses a single head. A dropout rate of 0.2 is applied, and the models are trained using Adam with a learning rate of $10^{-3}$, weight decay of $5\times10^{-4}$, batch size 32, and cross entropy loss. Training is performed for up to 300 epochs with an early stopping patience of 30 epochs.

The comparative models are evaluated using the same held out test partitions used for HGCN(O), enabling a direct comparison under identical test conditions. In addition, five fold cross validation is performed to assess the stability of their predictive performance without additional hyperparameter optimization. The folds follow the same dataset specific splitting principles used for HGCN(O): class stratification for Patients and sequence length stratification for BPI12-A and BPI12-O. Weighted F1 is used for the imbalanced Patients dataset, while accuracy is used for the balanced datasets.

\subsection{Evaluation Metrics}
\label{subsec:Eval}

Model performance is evaluated using accuracy, macro precision, macro recall, macro F1, weighted precision, weighted recall, and weighted F1. These metrics are reported consistently for HGCN(O), the comparative GNN baselines, and the ablation experiments. For the imbalanced Patients dataset, class specific precision, recall, and F1 scores are additionally reported to examine performance across individual outcome classes. For the BPI12-A and BPI12-O prefix experiments, classification results are reported at 30\%, 50\%, 70\%, and 100\% sequence prefixes. Cochran's $Q$ test is applied to the matched prediction outcomes at the event sequence level across the four prefix levels to determine whether the proportion of correctly classified event sequences differs significantly as additional sequence information becomes available. The corresponding $p$ values are reported together with the classification metrics.

\section{Results and Discussion}
\label{sec:RE}

This section presents and discusses the experimental results of HGCN(O), focusing on predictive performance, model behavior across datasets, the contribution of its main architectural components, and the analysis of computational complexity and scalability.

\subsection{Comparison with Attention Based GNNs}
\label{subsec:baseline_results}

Table~\ref{tab:gnn_baselines} compares the core T-GCN configurations with GAT and TransformerConv. On the balanced datasets, BPI12-A and BPI12-O, both T-GCNConv and T-GraphConv reach ceiling performance, while the two attention based models show lower and dataset dependent results. The Patients dataset provides a clearer comparison, with both T-GCN configurations outperforming GAT and TransformerConv in accuracy, macro F1, and weighted F1. The small difference between T-GCNConv and T-GraphConv further indicates that the observed performance is not tied to a particular graph convolution operator.

The comparison is not intended to establish the intrinsic superiority of GCNConv or GraphConv over attention based GNNs. Rather, it shows that the HGCN(O) framework, through automated configuration of its architecture and training strategy, can achieve strong predictive performance with comparatively simple graph convolution operators. This supports the role of automated hypermodel optimization within the integrated framework, while the ceiling performance on the balanced datasets is examined further through the prefix analysis.

\begin{table}[!htbp]
\centering
\caption{Performance Comparison with GNN Baselines}
\label{tab:gnn_baselines}
\setlength{\tabcolsep}{4pt}

\begin{tabular}{llccccccc}
\toprule
\multirow{2}{*}{\textbf{Dataset}} & \multirow{2}{*}{\textbf{Model}} & \multirow{2}{*}{\textbf{Accuracy}} & \multicolumn{3}{c}{\textbf{Macro Average}} & \multicolumn{3}{c}{\textbf{Weighted Average}} \\
\cmidrule(lr){4-6}\cmidrule(lr){7-9}
& & & \textbf{Precision} & \textbf{Recall} & \textbf{F1} & \textbf{Precision} & \textbf{Recall} & \textbf{F1} \\
\midrule
\multirow{4}{*}{BPI12-A} & T-GCNConv & 1.0000 & 1.0000 & 1.0000 & 1.0000 & 1.0000 & 1.0000 & 1.0000 \\
& T-GraphConv & 1.0000 & 1.0000 & 1.0000 & 1.0000 & 1.0000 & 1.0000 & 1.0000 \\
& GAT & 0.9610 & 0.9649 & 0.9612 & 0.9609 & 0.9651 & 0.9610 & 0.9609 \\
& TransformerConv & 0.9235 & 0.9338 & 0.9239 & 0.9228 & 0.9341 & 0.9235 & 0.9228 \\
\midrule
\multirow{4}{*}{BPI12-O} & T-GCNConv & 1.0000 & 1.0000 & 1.0000 & 1.0000 & 1.0000 & 1.0000 & 1.0000 \\
& T-GraphConv & 1.0000 & 1.0000 & 1.0000 & 1.0000 & 1.0000 & 1.0000 & 1.0000 \\
& GAT & 0.9501 & 0.9512 & 0.9504 & 0.9508 & 0.9502 & 0.9501 & 0.9501 \\
& TransformerConv & 0.9751 & 0.9779 & 0.9734 & 0.9752 & 0.9758 & 0.9751 & 0.9750 \\
\midrule
\multirow{4}{*}{Patients} & T-GCNConv & 0.8762 & 0.9436 & 0.9024 & 0.9084 & 0.8969 & 0.8762 & 0.8639 \\
& T-GraphConv & 0.8738 & 0.9343 & 0.9071 & 0.9044 & 0.8968 & 0.8738 & 0.8599 \\
& GAT & 0.8458 & 0.8762 & 0.8491 & 0.8392 & 0.8796 & 0.8458 & 0.8318 \\
& TransformerConv & 0.8575 & 0.8740 & 0.8698 & 0.8458 & 0.8821 & 0.8575 & 0.8453 \\
\bottomrule
\end{tabular}
\end{table}

\subsection{Evaluation under Class Imbalance}
\label{subsec:Imresult}

To examine how HGCN(O) adapts to class imbalance, the complete model family is evaluated on the Patients dataset using eight configurations formed by T-GCN, O-GCN, TP-GCN, and TE-GCN with GCNConv and GraphConv. Table~\ref{tab:confusion} reports the class specific precision, recall, and F1 scores together with the loss standard deviation during the 300 epoch retraining phase for the configurations selected through cross validation. Figure~\ref{fig:lcIm} presents the corresponding learning curves. The selected configurations vary across architectures and graph operators in their structural, regularization, optimizer, and scheduler settings. The complete hyperparameter configurations are reported in~\ref{sec:appendix}, Table~\ref{app_tab:Imresults}.

\begin{table}[!htbp]
\centering
\caption{Class Specific Performance and Training Stability of HGCN(O) Models on the Imbalanced Patients Dataset}
\label{tab:confusion}
\setlength{\tabcolsep}{2pt}
%========================
% Block 1
%========================
\begin{tabular}{c ccc ccc ccc ccc c}
\toprule
 & \multicolumn{3}{c}{\textbf{O-GCNConv}}
  & \multicolumn{3}{c}{\textbf{O-GraphConv}}
  & \multicolumn{3}{c}{\textbf{T-GCNConv}}
  & \multicolumn{3}{c}{\textbf{T-GraphConv}}
  &  \\
\cmidrule(lr){2-4}\cmidrule(lr){5-7}\cmidrule(lr){8-10}\cmidrule(lr){11-13}
Class & Precision & Recall & F1 & Precision & Recall & F1 & Precision & Recall & F1 & Precision & Recall & F1 & Support \\
\midrule
0 & \cellcolor[rgb]{ .792,  .929,  .984}1.0000      & \cellcolor[rgb]{ .969,  .78,  .675}1.0000      & \cellcolor[rgb]{ .8,  1,  .6}1.0000
  & \cellcolor[rgb]{ .792,  .929,  .984}1.0000      & \cellcolor[rgb]{ .969,  .78,  .675}1.0000      & \cellcolor[rgb]{ .8,  1,  .6}1.0000
  & \cellcolor[rgb]{ .792,  .929,  .984}1.0000      & \cellcolor[rgb]{ .969,  .78,  .675}1.0000       & \cellcolor[rgb]{ .8,  1,  .6}1.0000
  & \cellcolor[rgb]{ .792,  .929,  .984}1.0000      & \cellcolor[rgb]{ .969,  .78,  .675}1.0000       & \cellcolor[rgb]{ .8,  1,  .6}1.0000
  & 92 \\

1 & \cellcolor[rgb]{ .792,  .929,  .984}0.7838 & \cellcolor[rgb]{ .969,  .78,  .675}1.0000       & \cellcolor[rgb]{ .8,  1,  .6}0.8788
  & \cellcolor[rgb]{ .792,  .929,  .984}0.7953 & \cellcolor[rgb]{ .969,  .78,  .675}0.9828 & \cellcolor[rgb]{ .8,  1,  .6}0.8792
  & \cellcolor[rgb]{ .792,  .929,  .984}0.7838 & \cellcolor[rgb]{ .969,  .78,  .675}1.0000       & \cellcolor[rgb]{ .8,  1,  .6}0.8788
  & \cellcolor[rgb]{ .792,  .929,  .984}0.7838 & \cellcolor[rgb]{ .969,  .78,  .675}1.0000       & \cellcolor[rgb]{ .8,  1,  .6}0.8788
  & 174 \\

2 & \cellcolor[rgb]{ .792,  .929,  .984}1.0000      & \cellcolor[rgb]{ .969,  .78,  .675}1.0000       & \cellcolor[rgb]{ .8,  1,  .6}1.0000
  & \cellcolor[rgb]{ .792,  .929,  .984}0.625  & \cellcolor[rgb]{ .969,  .78,  .675}1.0000      & \cellcolor[rgb]{ .8,  1,  .6}0.7692
  & \cellcolor[rgb]{ .792,  .929,  .984}1.0000      & \cellcolor[rgb]{ .969,  .78,  .675}1.0000      & \cellcolor[rgb]{ .8,  1,  .6}1.0000
  & \cellcolor[rgb]{ .792,  .929,  .984}1.0000      & \cellcolor[rgb]{ .969,  .78,  .675}1.0000       & \cellcolor[rgb]{ .8,  1,  .6}1.0000
  & 5 \\

3 & \cellcolor[rgb]{ .792,  .929,  .984}0.9545 & \cellcolor[rgb]{ .969,  .78,  .675}1.0000      & \cellcolor[rgb]{ .8,  1,  .6}0.9767
  & \cellcolor[rgb]{ .792,  .929,  .984}1.0000      & \cellcolor[rgb]{ .969,  .78,  .675}0.9048 & \cellcolor[rgb]{ .8,  1,  .6}0.95
  & \cellcolor[rgb]{ .792,  .929,  .984}1.0000      & \cellcolor[rgb]{ .969,  .78,  .675}0.9048 & \cellcolor[rgb]{ .8,  1,  .6}0.95
  & \cellcolor[rgb]{ .792,  .929,  .984}0.9524 & \cellcolor[rgb]{ .969,  .78,  .675}0.9524 & \cellcolor[rgb]{ .8,  1,  .6}0.9524
  & 21 \\

4 & \cellcolor[rgb]{ .792,  .929,  .984}0.8649 & \cellcolor[rgb]{ .969,  .78,  .675}1.0000      & \cellcolor[rgb]{ .8,  1,  .6}0.9275
  & \cellcolor[rgb]{ .792,  .929,  .984}0.8611 & \cellcolor[rgb]{ .969,  .78,  .675}0.9688 & \cellcolor[rgb]{ .8,  1,  .6}0.9118
  & \cellcolor[rgb]{ .792,  .929,  .984}0.9143 & \cellcolor[rgb]{ .969,  .78,  .675}1.0000      & \cellcolor[rgb]{ .8,  1,  .6}0.9552
  & \cellcolor[rgb]{ .792,  .929,  .984}0.8889 & \cellcolor[rgb]{ .969,  .78,  .675}1.0000      & \cellcolor[rgb]{ .8,  1,  .6}0.9412
  & 32 \\

5 & \cellcolor[rgb]{ .792,  .929,  .984}1.0000      & \cellcolor[rgb]{ .969,  .78,  .675}0.4808 & \cellcolor[rgb]{ .8,  1,  .6}0.6494
  & \cellcolor[rgb]{ .792,  .929,  .984}0.931  & \cellcolor[rgb]{ .969,  .78,  .675}0.5192 & \cellcolor[rgb]{ .8,  1,  .6}0.6667
  & \cellcolor[rgb]{ .792,  .929,  .984}0.9636 & \cellcolor[rgb]{ .969,  .78,  .675}0.5096 & \cellcolor[rgb]{ .8,  1,  .6}0.6667
  & \cellcolor[rgb]{ .792,  .929,  .984}0.9808 & \cellcolor[rgb]{ .969,  .78,  .675}0.4904 & \cellcolor[rgb]{ .8,  1,  .6}0.6538
  & 104 \\

%Accuracy & \cellcolor[rgb]{ .792,  .929,  .984}       & \cellcolor[rgb]{ .969,  .78,  .675}       & \cellcolor[rgb]{ .8,  1,  .6}0.8738   & \cellcolor[rgb]{ .792,  .929,  .984}       & \cellcolor[rgb]{ .969,  .78,  .675}       & \cellcolor[rgb]{ .8,  1,  .6}0.8692  & \cellcolor[rgb]{ .792,  .929,  .984}       & \cellcolor[rgb]{ .969,  .78,  .675}       & \cellcolor[rgb]{ .8,  1,  .6}0.8762  & \cellcolor[rgb]{ .792,  .929,  .984}       & \cellcolor[rgb]{ .969,  .78,  .675}       & \cellcolor[rgb]{ .8,  1,  .6}0.8738  & 428 \\

Macro & \cellcolor[rgb]{ .792,  .929,  .984}0.9339 & \cellcolor[rgb]{ .969,  .78,  .675}0.9135 & \cellcolor[rgb]{ .8,  1,  .6}0.9054
  & \cellcolor[rgb]{ .792,  .929,  .984}0.8687 & \cellcolor[rgb]{ .969,  .78,  .675}0.8959 & \cellcolor[rgb]{ .8,  1,  .6}0.8628
  & \cellcolor[rgb]{ .792,  .929,  .984}0.9436 & \cellcolor[rgb]{ .969,  .78,  .675}0.9024 & \cellcolor[rgb]{ .8,  1,  .6}0.9084
  & \cellcolor[rgb]{ .792,  .929,  .984}0.9343 & \cellcolor[rgb]{ .969,  .78,  .675}0.9071 & \cellcolor[rgb]{ .8,  1,  .6}0.9044
  & 428 \\

Weighted & \cellcolor[rgb]{ .792,  .929,  .984}0.8998 & \cellcolor[rgb]{ .969,  .78,  .675}0.8738 & \cellcolor[rgb]{ .8,  1,  .6}0.859
  & \cellcolor[rgb]{ .792,  .929,  .984}0.8853 & \cellcolor[rgb]{ .969,  .78,  .675}0.8692 & \cellcolor[rgb]{ .8,  1,  .6}0.8581
  & \cellcolor[rgb]{ .792,  .929,  .984}0.8969 & \cellcolor[rgb]{ .969,  .78,  .675}0.8762 & \cellcolor[rgb]{ .8,  1,  .6}0.8639
  & \cellcolor[rgb]{ .792,  .929,  .984}0.8968 & \cellcolor[rgb]{ .969,  .78,  .675}0.8738 & \cellcolor[rgb]{ .8,  1,  .6}0.8599
  & 428 \\ 
\midrule
Accuracy & \multicolumn{3}{c}{0.8738} & \multicolumn{3}{c}{0.8692} & \multicolumn{3}{c}{0.8762} & \multicolumn{3}{c}{0.8738} & 428 \\
\cmidrule(lr){2-4}\cmidrule(lr){5-7}\cmidrule(lr){8-10}\cmidrule(lr){11-13}
Loss Std & \multicolumn{3}{c}{0.0103} & \multicolumn{3}{c}{0.0376} & \multicolumn{3}{c}{0.0395} & \multicolumn{3}{c}{0.0616} &  \\
\bottomrule
\end{tabular}

\vspace{0.7em}

%========================
% Block 2
%========================
\begin{tabular}{c ccc ccc ccc ccc c}
\toprule
  & \multicolumn{3}{c}{\textbf{TP-GCNConv}}
  & \multicolumn{3}{c}{\textbf{TP-GraphConv}}
  & \multicolumn{3}{c}{\textbf{TE-GCNConv}}
  & \multicolumn{3}{c}{\textbf{TE-GraphConv}}
  &  \\
\cmidrule(lr){2-4}\cmidrule(lr){5-7}\cmidrule(lr){8-10}\cmidrule(lr){11-13}
Class & Precision & Recall & F1 & Precision & Recall & F1 & Precision & Recall & F1 & Precision & Recall & F1 & Support \\
\midrule
0 & \cellcolor[rgb]{ .792,  .929,  .984}1.0000      & \cellcolor[rgb]{ .969,  .78,  .675}1.0000      & \cellcolor[rgb]{ .8,  1,  .6}1.0000
  & \cellcolor[rgb]{ .792,  .929,  .984}1.0000      & \cellcolor[rgb]{ .969,  .78,  .675}1.0000      & \cellcolor[rgb]{ .8,  1,  .6}1.0000
  & \cellcolor[rgb]{ .792,  .929,  .984}1.0000      & \cellcolor[rgb]{ .969,  .78,  .675}1.0000      & \cellcolor[rgb]{ .8,  1,  .6}1.0000
  & \cellcolor[rgb]{ .792,  .929,  .984}1.0000      & \cellcolor[rgb]{ .969,  .78,  .675}1.0000      & \cellcolor[rgb]{ .8,  1,  .6}1.0000
  & 92 \\

1 & \cellcolor[rgb]{ .792,  .929,  .984}0.7793 & \cellcolor[rgb]{ .969,  .78,  .675}0.9943 & \cellcolor[rgb]{ .8,  1,  .6}0.8737
  & \cellcolor[rgb]{ .792,  .929,  .984}0.7768 & \cellcolor[rgb]{ .969,  .78,  .675}1.0000      & \cellcolor[rgb]{ .8,  1,  .6}0.8744
  & \cellcolor[rgb]{ .792,  .929,  .984}0.7838 & \cellcolor[rgb]{ .969,  .78,  .675}1.0000      & \cellcolor[rgb]{ .8,  1,  .6}0.8788
  & \cellcolor[rgb]{ .792,  .929,  .984}0.7838 & \cellcolor[rgb]{ .969,  .78,  .675}1.0000      & \cellcolor[rgb]{ .8,  1,  .6}0.8788
  & 174 \\

2 & \cellcolor[rgb]{ .792,  .929,  .984}1.0000      & \cellcolor[rgb]{ .969,  .78,  .675}1.0000      & \cellcolor[rgb]{ .8,  1,  .6}1.0000
  & \cellcolor[rgb]{ .792,  .929,  .984}1.0000      & \cellcolor[rgb]{ .969,  .78,  .675}1.0000      & \cellcolor[rgb]{ .8,  1,  .6}1.0000
  & \cellcolor[rgb]{ .792,  .929,  .984}1.0000      & \cellcolor[rgb]{ .969,  .78,  .675}1.0000      & \cellcolor[rgb]{ .8,  1,  .6}1.0000
  & \cellcolor[rgb]{ .792,  .929,  .984}1.0000      & \cellcolor[rgb]{ .969,  .78,  .675}1.0000      & \cellcolor[rgb]{ .8,  1,  .6}1.0000
  & 5 \\

3 & \cellcolor[rgb]{ .792,  .929,  .984}0.9091 & \cellcolor[rgb]{ .969,  .78,  .675}0.9524 & \cellcolor[rgb]{ .8,  1,  .6}0.9302
  & \cellcolor[rgb]{ .792,  .929,  .984}1.0000      & \cellcolor[rgb]{ .969,  .78,  .675}1.0000      & \cellcolor[rgb]{ .8,  1,  .6}1.0000
  & \cellcolor[rgb]{ .792,  .929,  .984}1.0000      & \cellcolor[rgb]{ .969,  .78,  .675}1.0000      & \cellcolor[rgb]{ .8,  1,  .6}1.0000
  & \cellcolor[rgb]{ .792,  .929,  .984}1.0000      & \cellcolor[rgb]{ .969,  .78,  .675}0.9524 & \cellcolor[rgb]{ .8,  1,  .6}0.9756
  & 21 \\

4 & \cellcolor[rgb]{ .792,  .929,  .984}0.9091 & \cellcolor[rgb]{ .969,  .78,  .675}0.9375 & \cellcolor[rgb]{ .8,  1,  .6}0.9231
  & \cellcolor[rgb]{ .792,  .929,  .984}0.9333 & \cellcolor[rgb]{ .969,  .78,  .675}0.875  & \cellcolor[rgb]{ .8,  1,  .6}0.9032
  & \cellcolor[rgb]{ .792,  .929,  .984}0.8649 & \cellcolor[rgb]{ .969,  .78,  .675}1.0000      & \cellcolor[rgb]{ .8,  1,  .6}0.9275
  & \cellcolor[rgb]{ .792,  .929,  .984}0.9143 & \cellcolor[rgb]{ .969,  .78,  .675}1.0000      & \cellcolor[rgb]{ .8,  1,  .6}0.9552
  & 32 \\

5 & \cellcolor[rgb]{ .792,  .929,  .984}0.9815 & \cellcolor[rgb]{ .969,  .78,  .675}0.5096 & \cellcolor[rgb]{ .8,  1,  .6}0.6709
  & \cellcolor[rgb]{ .792,  .929,  .984}0.9643 & \cellcolor[rgb]{ .969,  .78,  .675}0.5192 & \cellcolor[rgb]{ .8,  1,  .6}0.675
  & \cellcolor[rgb]{ .792,  .929,  .984}1.0000      & \cellcolor[rgb]{ .969,  .78,  .675}0.4904 & \cellcolor[rgb]{ .8,  1,  .6}0.6581
  & \cellcolor[rgb]{ .792,  .929,  .984}0.9815 & \cellcolor[rgb]{ .969,  .78,  .675}0.5096 & \cellcolor[rgb]{ .8,  1,  .6}0.6709
  & 104 \\

%Accuracy & \cellcolor[rgb]{ .792,  .929,  .984}       & \cellcolor[rgb]{ .969,  .78,  .675}       & \cellcolor[rgb]{ .8,  1,  .6}0.8715   & \cellcolor[rgb]{ .792,  .929,  .984}       & \cellcolor[rgb]{ .969,  .78,  .675}       & \cellcolor[rgb]{ .8,  1,  .6}0.8738  & \cellcolor[rgb]{ .792,  .929,  .984}       & \cellcolor[rgb]{ .969,  .78,  .675}       & \cellcolor[rgb]{ .8,  1,  .6}0.8762  & \cellcolor[rgb]{ .792,  .929,  .984}       & \cellcolor[rgb]{ .969,  .78,  .675}       & \cellcolor[rgb]{ .8,  1,  .6}0.8785  & 428 \\

Macro & \cellcolor[rgb]{ .792,  .929,  .984}0.9298 & \cellcolor[rgb]{ .969,  .78,  .675}0.899  & \cellcolor[rgb]{ .8,  1,  .6}0.8997
  & \cellcolor[rgb]{ .792,  .929,  .984}0.9457 & \cellcolor[rgb]{ .969,  .78,  .675}0.899  & \cellcolor[rgb]{ .8,  1,  .6}0.9088
  & \cellcolor[rgb]{ .792,  .929,  .984}0.9414 & \cellcolor[rgb]{ .969,  .78,  .675}0.9151 & \cellcolor[rgb]{ .8,  1,  .6}0.9107
  & \cellcolor[rgb]{ .792,  .929,  .984}0.9466 & \cellcolor[rgb]{ .969,  .78,  .675}0.9103 & \cellcolor[rgb]{ .8,  1,  .6}0.9134
  & 428 \\

Weighted & \cellcolor[rgb]{ .792,  .929,  .984}0.8945 & \cellcolor[rgb]{ .969,  .78,  .675}0.8715 & \cellcolor[rgb]{ .8,  1,  .6}0.8595
  & \cellcolor[rgb]{ .792,  .929,  .984}0.8956 & \cellcolor[rgb]{ .969,  .78,  .675}0.8738 & \cellcolor[rgb]{ .8,  1,  .6}0.8627
  & \cellcolor[rgb]{ .792,  .929,  .984}0.902  & \cellcolor[rgb]{ .969,  .78,  .675}0.8762 & \cellcolor[rgb]{ .8,  1,  .6}0.8622
  & \cellcolor[rgb]{ .792,  .929,  .984}0.9012 & \cellcolor[rgb]{ .969,  .78,  .675}0.8785 & \cellcolor[rgb]{ .8,  1,  .6}0.8662
  & 428 \\
\midrule
Accuracy & \multicolumn{3}{c}{0.8715} & \multicolumn{3}{c}{0.8738} & \multicolumn{3}{c}{0.8762} & \multicolumn{3}{c}{0.8785} & 428 \\
\cmidrule(lr){2-4}\cmidrule(lr){5-7}\cmidrule(lr){8-10}\cmidrule(lr){11-13}
Loss Std & \multicolumn{3}{c}{0.0526} & \multicolumn{3}{c}{0.0564} & \multicolumn{3}{c}{0.0260} & \multicolumn{3}{c}{0.0467} &  \\

\bottomrule
\end{tabular}
\end{table}

\subsubsection{Overall Performance}
\label{sbsbsec:IMper}

Across the eight HGCN(O) configurations, overall accuracy remains relatively close, while the class specific metrics show clearer differences under the strong imbalance of the Patients dataset. The two level T-GCN architecture generally improves upon the one level O-GCN reference, supporting the separation of node level and graph level information in the core model. The extension models further show that auxiliary duration and activity representations can improve the learned representation, although the magnitude of this effect depends on the graph convolution operator. GCNConv performs strongly with the core two level structure, whereas GraphConv benefits more noticeably from the additional representations.

This difference is also reflected in training stability. GCNConv models generally exhibit lower loss variation and smoother learning curves, with O-GCNConv recording the lowest loss standard deviation despite its simpler input structure. GraphConv shows greater variation in the core configuration, but this is reduced when auxiliary representations are introduced, accompanied by corresponding improvements in F1 performance. The results therefore suggest that the two graph operators respond differently to the input representation: GCNConv remains effective with a comparatively compact representation, while GraphConv gains more from enriched node level representations.

\begin{figure}[pos = htbp]
\centering
\includegraphics[width=\textwidth]{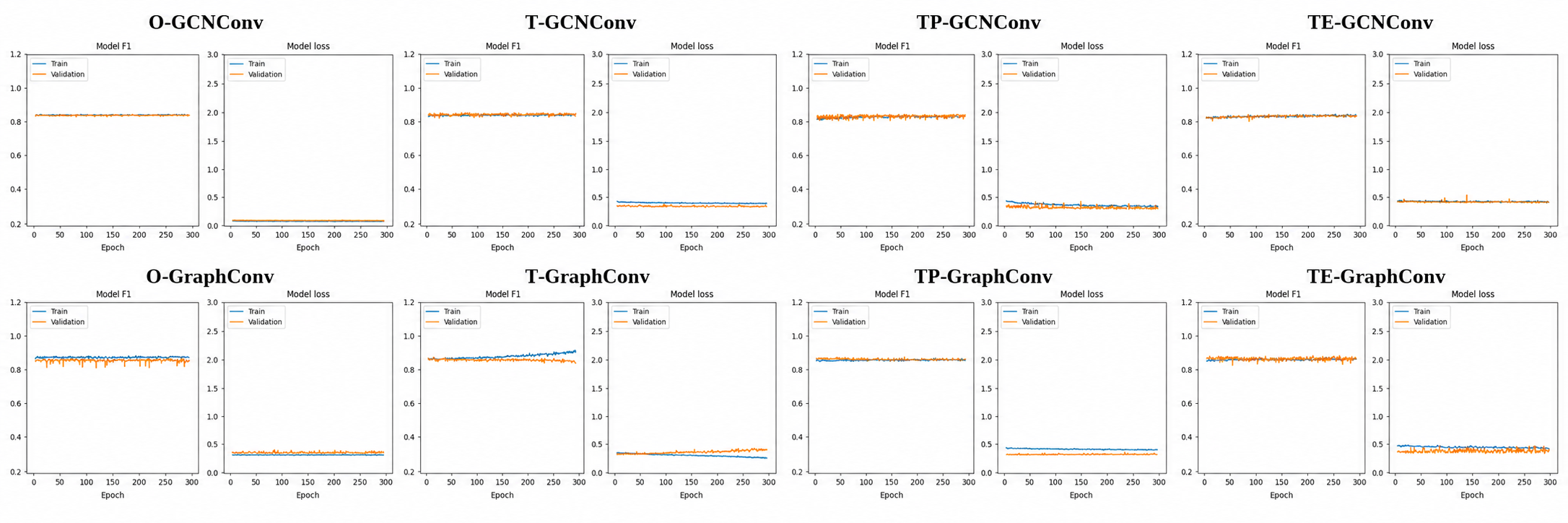}
\caption{Training and test learning curves for the eight HGCN(O) configurations on the imbalanced Patients dataset over 300 epochs. The upper row shows O-GCNConv, T-GCNConv, TP-GCNConv, and TE-GCNConv, while the lower row shows the corresponding GraphConv configurations. For each model, the left panel reports the F1 score and the right panel reports the loss for the training and test sets.}
\label{fig:lcIm}
\end{figure}

The learning curves in Figure~\ref{fig:lcIm} show stable convergence for most configurations, although some models exhibit larger fluctuations under the imbalanced class distribution. Training and test losses remain close in most cases, indicating limited divergence during the final training stage. Early stopping during hyperparameter optimization identifies the epoch with the best validation performance, while F1 varies only modestly during the 300 epoch retraining phase, typically within 1 to 3\%. This variation is consistent with the stochastic nature of model retraining rather than a systematic loss of generalization.

\subsubsection{Class Specific Performance}
\label{sbsbsec:clsperf}

Across all models, recall for classes 0 through 4 is at least 0.875, with approximately 75\% of the corresponding results achieving perfect recall. These results indicate that the five classes are identified reliably across the evaluated configurations. In contrast, class 5 exhibits consistently lower recall of approximately 0.5, with noticeable variation among model configurations. The prediction errors suggest that this class is less separable under the learned representations, particularly in relation to classes 1 and 4.

Among the six classes, class 0 achieves perfect precision, recall, and F1 across all models. Class 1 shows only minor variation among models, with precision concentrated around 0.78, while recall remains above 0.98 for most configurations and reaches 1.0 in several cases. Analysis of the confusion matrices shows that approximately 44 to 48 instances of class 5 are consistently misclassified as class 1 across the eight models. These errors contribute directly to the lower precision of class 1 and indicate substantial overlap between the learned representations of the two classes.

Class 2 presents a different challenge because it contains only five instances in the test set. Such a small sample size makes its class specific performance particularly sensitive to individual classification errors. Although most models achieve perfect scores for this class, O-GraphConv shows lower performance, indicating greater sensitivity to this highly underrepresented class.

Despite having fewer samples than classes 0 and 1, classes 3 and 4 generally achieve satisfactory performance, with several models attaining perfect recall. Analysis of the confusion matrices further shows that approximately three to five instances of class 5 are misclassified as class 4, contributing to reduced precision for class 4. These results show that the main classification difficulty is concentrated in class 5 rather than uniformly distributed across the minority classes. In applications where minority outcomes carry greater decision importance, model optimization may therefore require an explicit trade off between overall predictive performance and sensitivity to difficult classes.

\subsubsection{Hyperparameter Configurations}
\label{sbsbsec:IMhyper}

The optimized configurations reported in~\ref{sec:appendix}, Table~\ref{app_tab:Imresults}, show substantial variation across the eight HGCN(O) models. Differences are observed in network depth and width, optimizer and learning rate settings, regularization, batch size, activation functions, and the use of skip connections. No single configuration is consistently preferred across architectures or graph convolution operators. In particular, stronger performance is not systematically associated with deeper or wider networks; comparatively compact configurations can perform as well as, or better than, more complex alternatives. This variation supports the use of architecture specific optimization rather than applying a common manually selected configuration to the complete model family.

The selected training configurations also differ between GCNConv and GraphConv models, indicating that the two operators respond differently to optimization choices. Adam and RMSprop are both represented among the selected configurations, while learning rate, weight decay, batch size, and regularization vary across models. Skip connections are selected in several better performing configurations but are not universally required, further indicating that their contribution depends on the surrounding architecture. The optimized configurations show that performance depends on the joint selection of architectural and training parameters rather than on any individual hyperparameter. This reflects the adaptive role of the HGCN(O) hypermodel, which configures each architecture according to the data, input representation, and graph operator within a common automated search framework.

\subsection{Evaluation on Balanced Datasets}
\label{subsec:bpiresult}

\begin{figure}[pos = htbp]
\centering
\includegraphics[width=\textwidth]{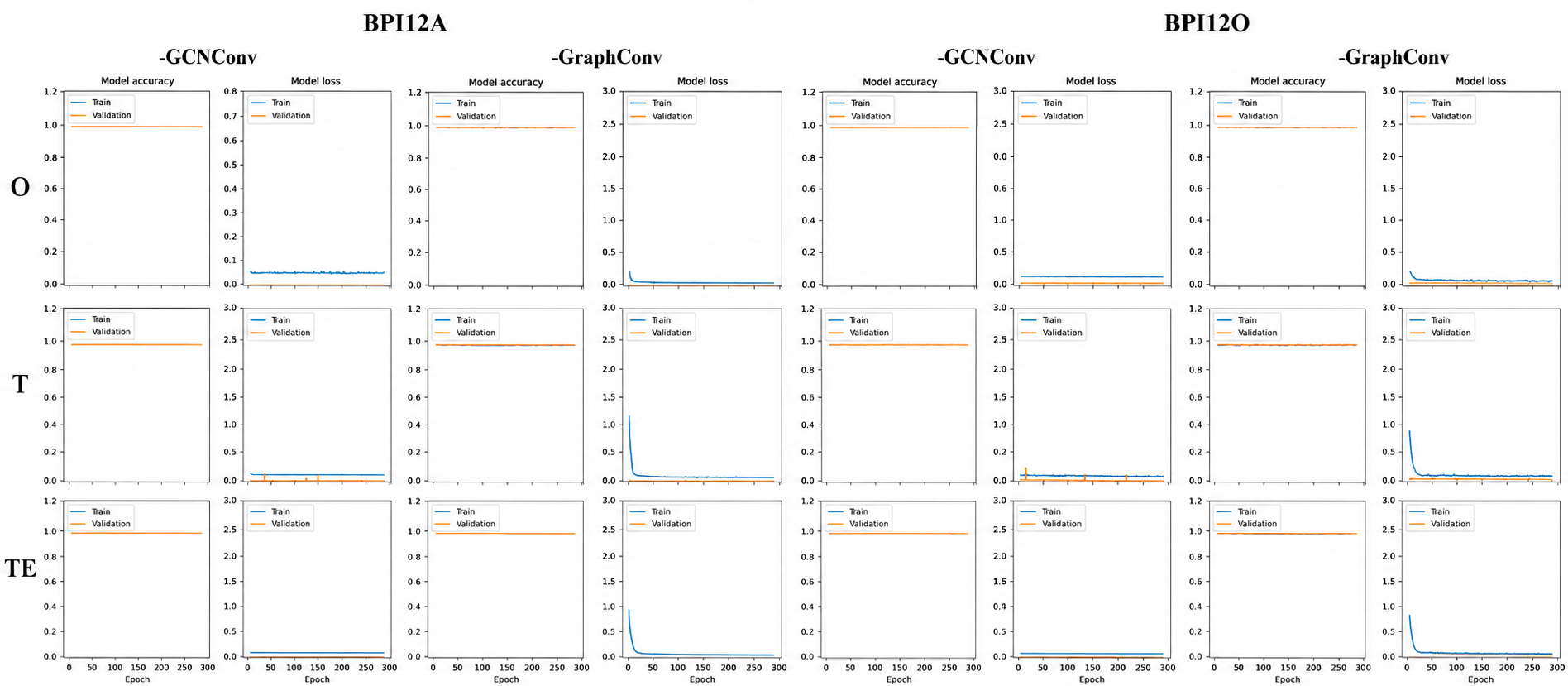}
\caption{Training and test learning curves for the selected HGCN(O) configurations on the balanced BPI12-A and BPI12-O datasets over 300 epochs. Rows correspond to the O-GCN, T-GCN, and TE-GCN architectures, while columns compare their GCNConv and GraphConv implementations. For each configuration, the left panel reports accuracy and the right panel reports loss on the training and test sets.}
\label{fig:lcBP}
\end{figure}

For the balanced BPI12-A and BPI12-O datasets, we evaluate the T-GCN core model, the O-GCN one level reference model, and the TE-GCN extension with both GCNConv and GraphConv. Since activity durations are zero for these datasets, the pseudo informative duration embedding is not applicable, and TE-GCN directly incorporates the learned activity embedding into the T-GCN architecture. The optimized hyperparameter configurations are reported in~\ref{sec:appendix}, Tables~\ref{app_tab:BPIResultsA} and~\ref{app_tab:BPIResultsO}. Figure~\ref{fig:lcBP} presents the final retraining learning curves for the model configuration selected through cross validation for each architecture and graph operator.

The classification reports are not included since all six HGCN(O) configurations achieve precision, recall, and F1 scores of 1.0 across all classes, making the reports redundant. The GCNConv models generally exhibit greater training stability than the GraphConv models, consistent with the observations on the imbalanced Patients dataset. Several models also converge rapidly, reaching their best validation performance within 13 to 31 epochs during hyperparameter optimization. Although all six configurations achieve perfect accuracy on the balanced datasets, their optimized configurations differ substantially in architecture and training strategy. Since predictive performance at sequence completion does not distinguish the models, these configurations provide additional insight into how the HGCN(O) hypermodel adapts each architecture to the characteristics of the dataset.

\subsubsection{Hyperparameter Configurations}
\label{sbsbse:blhyper}

The optimized configurations reveal several structural patterns across the balanced datasets. In the graph convolution branch, GraphConv is generally associated with greater depth, whereas T-GCNConv and O-GCNConv attain the same predictive performance with shallower configurations. This pattern changes when the activity embedding is introduced in TE-GCN, indicating that the preferred graph depth depends jointly on the convolution operator and the node representation. Skip connections are selected mainly in GCNConv configurations, while \emph{LeakyReLU} and \emph{GELU} dominate the graph convolution layers. Dropout and batch normalization parameters, in contrast, vary considerably across models without a clear common pattern.

The dense branches show a different configuration profile. For T-GCN and O-GCN, GraphConv is generally paired with shallower dense structures than GCNConv, whereas this distinction becomes less evident in TE-GCN. \emph{ReLU} is the dominant activation in the dense layers, contrasting with the activation choices in the graph convolution branch. Differences are also observed between the two datasets: BPI12-O tends to select deeper graph and dense structures together with higher dropout rates than BPI12-A, despite both datasets reaching identical final classification performance. Thus, the hypermodel reaches the same predictive performance through substantially different architectural configurations.

The training parameters show similar variation. Adam is selected in most balanced dataset configurations, while RMSprop and SGD appear in specific models. No common preference emerges for the learning rate scheduler, and the selected learning rate, weight decay, and batch size differ across architectures. These patterns should be interpreted within the jointly optimized HGCN(O) configurations rather than as isolated effects of individual hyperparameters. 

\subsubsection{Prefix Based Early Prediction}
\label{sbsbsec:prefix}

Although the balanced datasets reach ceiling performance using complete sequences, this setting does not reveal how the models behave when only partial information is available. Predictive capability under incomplete information is important more generally and is particularly relevant to predictive process monitoring, where predictions are expected before process completion. For an event sequence $G_j$ with $n_j$ events, a prefix at completion level $\rho$ retains a proportion $\rho$ of the events from the beginning of the sequence and constructs the corresponding event graph using the same graph representation as the complete sequence. For each completion level $\rho\in\{30\%,50\%,70\%,100\%\}$, the core two level T-GCN model is trained and evaluated using event sequences truncated to the corresponding prefix. By progressively varying the amount of observed information, the experiment examines how prediction evolves as event sequence information becomes available, how sensitive performance is to sequence completeness, and whether the optimized HGCN(O) framework can provide reliable predictions before sequence completion. The results for BPI12-A and BPI12-O are reported in Table~\ref{tab:prefix_results}.

\begin{table}[!htbp]
\centering
\caption{Predictive Performance of T-GCNConv and T-GraphConv Across Sequence Completion Levels on the BPI12-A/O Datasets}
\label{tab:prefix_results}
\setlength{\tabcolsep}{3pt}

\begin{tabular}{lllccccccc}
\toprule
\multirow{2}{*}{\textbf{Dataset}} &
\multirow{2}{*}{\textbf{Model}} &
\multirow{2}{*}{\textbf{Prefix (\%)}} &
\multirow{2}{*}{\textbf{Accuracy}} &
\multicolumn{3}{c}{\textbf{Macro Average}} &
\multicolumn{3}{c}{\textbf{Weighted Average}} \\
\cmidrule(lr){5-7}\cmidrule(lr){8-10}
& & & &
\textbf{Precision} & \textbf{Recall} & \textbf{F1} &
\textbf{Precision} & \textbf{Recall} & \textbf{F1} \\
\midrule

\multirow{10}{*}{BPI12-A}
& \multirow{5}{*}{T-GCNConv}
& 30  & 0.7114 & 0.7121 & 0.7110 & 0.7094 & 0.7122 & 0.7114 & 0.7096 \\
& & 50  & 0.7286 & 0.7700 & 0.7279 & 0.7211 & 0.7704 & 0.7286 & 0.7217 \\
& & 70  & 0.8396 & 0.8513 & 0.8390 & 0.8367 & 0.8511 & 0.8396 & 0.8369 \\
& & 100 & 1.0000 & 1.0000 & 1.0000 & 1.0000 & 1.0000 & 1.0000 & 1.0000 \\
\cmidrule(lr){3-10}
& & \multicolumn{8}{l}{\textit{Cochran's $Q=509.3361$, $p=4.5221\times10^{-110}$ ($p<0.001$)}} \\
\cmidrule(lr){2-10}

& \multirow{5}{*}{T-GraphConv}
& 30  & 0.7114 & 0.7121 & 0.7110 & 0.7094 & 0.7122 & 0.7114 & 0.7096 \\
& & 50  & 0.7286 & 0.7780 & 0.7279 & 0.7192 & 0.7784 & 0.7286 & 0.7198 \\
& & 70  & 0.8381 & 0.8492 & 0.8376 & 0.8353 & 0.8490 & 0.8381 & 0.8354 \\
& & 100 & 1.0000 & 1.0000 & 1.0000 & 1.0000 & 1.0000 & 1.0000 & 1.0000 \\
\cmidrule(lr){3-10}
& & \multicolumn{8}{l}{\textit{Cochran's $Q=509.7034$, $p=3.7650\times10^{-110}$ ($p<0.001$)}} \\

\midrule

\multirow{10}{*}{BPI12-O}
& \multirow{5}{*}{T-GCNConv}
& 30  & 0.4657 & 0.4688 & 0.4426 & 0.4270 & 0.4681 & 0.4657 & 0.4381 \\
& & 50  & 0.7297 & 0.7478 & 0.7103 & 0.6978 & 0.7415 & 0.7297 & 0.7056 \\
& & 70  & 0.7630 & 0.7663 & 0.7631 & 0.7645 & 0.7662 & 0.7630 & 0.7645 \\
& & 100 & 1.0000 & 1.0000 & 1.0000 & 1.0000 & 1.0000 & 1.0000 & 1.0000 \\
\cmidrule(lr){3-10}
& & \multicolumn{8}{l}{\textit{Cochran's $Q=390.9097$, $p=2.9061\times10^{-84}$ ($p<0.001$)}} \\
\cmidrule(lr){2-10}

& \multirow{5}{*}{T-GraphConv}
& 30  & 0.4574 & 0.4856 & 0.4295 & 0.4030 & 0.4813 & 0.4574 & 0.4148 \\
& & 50  & 0.7193 & 0.7480 & 0.6981 & 0.6841 & 0.7404 & 0.7193 & 0.6924 \\
& & 70  & 0.7672 & 0.7720 & 0.7618 & 0.7643 & 0.7689 & 0.7672 & 0.7655 \\
& & 100 & 1.0000 & 1.0000 & 1.0000 & 1.0000 & 1.0000 & 1.0000 & 1.0000 \\
\cmidrule(lr){3-10}
& & \multicolumn{8}{l}{\textit{Cochran's $Q=404.4000$, $p=2.4664\times10^{-87}$ ($p<0.001$)}} \\

\bottomrule
\end{tabular}
\end{table}

Table~\ref{tab:prefix_results} shows that predictive performance increases with sequence completion on both datasets. For BPI12-A, accuracy rises from 0.7114 at 30\% to approximately 0.84 at 70\%, while macro and weighted F1 follow the same pattern. T-GCNConv and T-GraphConv produce almost identical results across all prefix levels. For BPI12-O, performance is more sensitive to the available sequence information: accuracy increases from approximately 0.46 at 30\% to approximately 0.72 at 50\% and 0.76 at 70\%. The corresponding macro and weighted F1 scores show a similar increase.

Cochran's $Q$ test confirms significant differences in prediction correctness across the four sequence completion levels for both T-GCN configurations on both datasets ($p<0.001$). The results indicate that the core T-GCN model can provide useful predictions before sequence completion, while the achievable performance depends on how much event sequence information is available.

\subsection{Ablation Analysis}
\label{subsec:ablation}

The ablation study evaluates the contribution of the structural components of T-GCN under automated hyperparameter optimization. Each ablated model is reoptimized using the same HGCN(O) hypermodel procedure. Two components are assessed. First, temporal edge weights $\mathbf{w}_{G_j}$ are removed while preserving the directed graph topology, isolating the contribution of the temporal information encoded between consecutive events. Second, the graph attribute vector $\mathbf{v}_{G_j}$ is removed from T-GCN, isolating the contribution of sequence level information incorporated at the graph level. 

The first ablation tests whether temporal edge weights enable the GNN to distinguish event sequences with identical event order but different temporal progression. Without these weights, such sequences induce the same graph connectivity despite potentially different time spans between events. The second ablation directly examines the two level design of T-GCN. Event attributes are processed through graph convolution at the node level, while sequence attributes $\mathbf{v}_{G_j}$ are incorporated after graph pooling, preserving the distinction between event level and graph level information. O-GCN provides an additional reference in Section~\ref{sbsbsec:IMper}: it retains the same sequence attributes but incorporates them at the node level. Its weaker performance on the imbalanced dataset suggests that the contribution of sequence information depends not only on its availability but also on where it is incorporated within the graph architecture. Table~\ref{tab:ablation} reports the effects of removing temporal edge weights and graph attributes from T-GCNConv and T-GraphConv across the three datasets.

On BPI12-A, removing either component reduces accuracy from 1.0 to approximately 0.96, with similar reductions in macro and weighted F1. BPI12-O shows the same pattern with a smaller decrease, with accuracy remaining around 0.98 after either ablation. The consistent decline across both graph convolution operators indicates that temporal edge information and graph attributes contribute to the complete T-GCN representation even when the full model reaches ceiling performance.

The Patients dataset provides a more discriminative assessment. For T-GCNConv, both ablations reduce accuracy and macro F1, although the effect is moderate. T-GraphConv is more sensitive to the removal of either component, particularly in macro F1, which decreases from 0.9044 to 0.8531 without temporal edge weights and to 0.8558 without graph attributes. Across the three datasets, the full architecture provides the strongest and most consistent performance. Overall, the ablation results indicate that the performance of HGCN(O) is supported by its integrated graph design, where temporal relations are encoded at the edge level and sequence attributes are incorporated at the graph level, rather than by automated hyperparameter optimization alone.

\begin{table}[!htbp]
\centering
\caption{Ablation Results of T-GCNConv and T-GraphConv on the Three Datasets}
\label{tab:ablation}

\setlength{\tabcolsep}{1.5pt}

\begin{tabular}{lccccccccccccccc}
\toprule
\multirow{3}{*}{\textbf{Dataset}} &
\multirow{3}{*}{\textbf{Model}} &
\multicolumn{7}{c}{\textbf{T-GCNConv}} &
\multicolumn{7}{c}{\textbf{T-GraphConv}} \\
\cmidrule(lr){3-9}\cmidrule(lr){10-16}

& &
\multirow{2}{*}{\textbf{Acc}} &
\multicolumn{3}{c}{\textbf{Macro Average}} &
\multicolumn{3}{c}{\textbf{Weighted Average}} &
\multirow{2}{*}{\textbf{Acc}} &
\multicolumn{3}{c}{\textbf{Macro Average}} &
\multicolumn{3}{c}{\textbf{Weighted Average}} \\
\cmidrule(lr){4-6}\cmidrule(lr){7-9}
\cmidrule(lr){11-13}\cmidrule(lr){14-16}

& &
& \textbf{P} & \textbf{R} & \textbf{F1}
& \textbf{P} & \textbf{R} & \textbf{F1}
&
& \textbf{P} & \textbf{R} & \textbf{F1}
& \textbf{P} & \textbf{R} & \textbf{F1} \\

\midrule

\multirow{3}{*}{BPI12-A}
& Full
& 1.0000 & 1.0000 & 1.0000 & 1.0000 & 1.0000 & 1.0000 & 1.0000
& 1.0000 & 1.0000 & 1.0000 & 1.0000 & 1.0000 & 1.0000 & 1.0000 \\

& $-\mathbf{w}_{G_j}$
& 0.9625 & 0.9661 & 0.9627 & 0.9624 & 0.9663 & 0.9625 & 0.9624
& 0.9633 & 0.9654 & 0.9634 & 0.9632 & 0.9656 & 0.9633 & 0.9632 \\

& $-\mathbf{v}_{G_j}$
& 0.9610 & 0.9649 & 0.9612 & 0.9609 & 0.9651 & 0.9610 & 0.9609
& 0.9610 & 0.9649 & 0.9612 & 0.9609 & 0.9651 & 0.9610 & 0.9609 \\

\midrule

\multirow{3}{*}{BPI12-O}
& Full
& 1.0000 & 1.0000 & 1.0000 & 1.0000 & 1.0000 & 1.0000 & 1.0000
& 1.0000 & 1.0000 & 1.0000 & 1.0000 & 1.0000 & 1.0000 & 1.0000 \\

& $-\mathbf{w}_{G_j}$
& 0.9813 & 0.9844 & 0.9792 & 0.9812 & 0.9822 & 0.9813 & 0.9812
& 0.9834 & 0.9856 & 0.9815 & 0.9831 & 0.9839 & 0.9834 & 0.9833 \\

& $-\mathbf{v}_{G_j}$
& 0.9813 & 0.9830 & 0.9802 & 0.9814 & 0.9816 & 0.9813 & 0.9812
& 0.9792 & 0.9827 & 0.9769 & 0.9791 & 0.9803 & 0.9792 & 0.9791 \\

\midrule

\multirow{3}{*}{Patients}
& Full
& 0.8762 & 0.9436 & 0.9024 & 0.9084 & 0.8969 & 0.8762 & 0.8639
& 0.8738 & 0.9343 & 0.9071 & 0.9044 & 0.8968 & 0.8738 & 0.8599 \\

& $-\mathbf{w}_{G_j}$
& 0.8638 & 0.9302 & 0.8999 & 0.8977 & 0.9003 & 0.8638 & 0.8501
& 0.8592 & 0.8761 & 0.8602 & 0.8531 & 0.8724 & 0.8592 & 0.8434 \\

& $-\mathbf{v}_{G_j}$
& 0.8662 & 0.9328 & 0.8979 & 0.9014 & 0.8965 & 0.8662 & 0.8540
& 0.8585 & 0.8961 & 0.8737 & 0.8558 & 0.8807 & 0.8585 & 0.8471 \\

\bottomrule
\multicolumn{16}{l}{* $-\mathbf{w}_{G_j}$: temporal edge weights removed; $-\mathbf{v}_{G_j}$: graph attributes removed. Acc: Accuracy; P: Precision; R: Recall.}\\
\end{tabular}
\end{table}

\subsection{Computational Complexity and Scalability}
\label{subsec:complexity}

\subsubsection{Computational Complexity}
\label{sbsbsec:comput}

Beyond predictive performance, computational efficiency and scalability are important for assessing the practical applicability of the evaluated models to event sequence data. In HGCN(O), each event sequence $G_j$ is represented as a sparse directed graph in which each event is connected only to its immediate successor. A sequence containing $n_j$ events therefore has $m_j=n_j-1$ directed edges, so both the number of nodes and edges increase linearly with sequence length.

For a graph convolution layer $l$ with input dimension $d_{l-1}$ and output dimension $d_l$, the dominant computational cost of GCNConv is $\mathcal{O}(n_jd_{l-1}d_l+m_jd_l)$, where the two terms correspond to node feature transformation and neighborhood aggregation, respectively. GraphConv has comparable complexity, $\mathcal{O}(n_jd_{l-1}d_l+m_jd_{l-1})$, as neighborhood information is aggregated before the separate transformations of the central node and neighboring representations. Since $m_j=n_j-1$, both operators scale linearly with sequence length when the feature dimensions are fixed. Graph pooling is also linear in $n_j$, whereas the fully connected branch for graph attributes is independent of sequence length.

The architectural variants introduce different computational overheads without changing this scaling behavior. O-GCN increases the input feature dimension by incorporating graph attributes at the node level, while TP-GCN and TE-GCN introduce additional graph processing branches for their auxiliary representations. These components increase the computation associated with each event sequence but preserve the linear dependence on sequence length.

At the dataset level, let $S$ denote the number of event sequences and $N=\sum_{j=1}^{S}n_j$ the total number of events. The corresponding graph representation contains $N-S$ directed edges. For fixed architectural dimensions, the graph processing cost therefore grows approximately linearly with the total number of events. Memory requirements follow the same sparse structure because node representations and observed edges are stored directly without constructing dense $n_j\times n_j$ adjacency matrices.

Automated hypermodel optimization introduces additional computational cost during model development because multiple architecture and training configurations are evaluated. For $R$ optimization trials, the total search cost depends on the accumulated training cost of the evaluated configurations. Early stopping and Optuna pruning reduce this cost by terminating trials before completion of the full training schedule when appropriate. Once the selected configuration is obtained, the optimization procedure introduces no additional computation during inference.

\subsubsection{Alternative GNN Operators}
\label{sbsbsec:alterGNNs}

GAT and TransformerConv operate on the same sparse event sequence graphs and restrict attention to the observed edges. For fixed hidden dimensions and numbers of attention heads, their graph processing cost therefore also grows linearly with the number of edges. Attention based message passing, however, introduces additional feature projections and attention coefficient computations for connected nodes. The comparison therefore evaluates whether HGCN(O) can achieve strong predictive performance using comparatively simple graph convolution operators rather than establishing the intrinsic superiority of one operator family over another.

GIN was not included as an additional baseline because its primary strength lies in discriminative neighborhood aggregation over graph topology, whereas the topology of the event sequence graphs is deliberately constrained by consecutive event order. Moreover, standard GIN does not directly incorporate the continuous temporal edge weights used in the proposed representation. An edge aware extension such as GINE would require a different treatment of edge information and would therefore not constitute a direct operator replacement under the same input formulation. GAT and TransformerConv were selected instead as alternative message passing operators that can operate on the same directed event graph and incorporate its temporal edge information. Future work may examine additional operators within the two level T-GCN structure to assess their interaction with the proposed representation.

\subsubsection{Scalability and Practical Considerations}
\label{sbsbsec:scal}

For the three academic benchmark datasets considered in this study, the selected model configurations completed the final training stage within approximately one hour using the limited computational resources described in Section~\ref{sec:Exp}. This indicates that the current HGCN(O) implementation is computationally manageable at the scale evaluated here. The sparse event graph representation also avoids the memory cost associated with dense adjacency structures, while the underlying graph operations retain linear dependence on the number of observed events and edges for fixed architectural dimensions.

Scalability to substantially larger industrial event sequence data, however, introduces a different challenge. Although the graph representation itself remains sparse, extensive hyperparameter search can become increasingly expensive as the number and length of event sequences grow. Practical deployment at this scale therefore requires balancing computational cost against the extent of hyperparameter search. This issue becomes more important if attention based operators are incorporated into the same framework, since operator choice would then affect both predictive performance and computational demand. Evaluating this balance across graph convolution and attention based operators on large industrial datasets is an important direction for future work.

\section{Conclusion}
\label{sec:conclu}

This study developed and evaluated HGCN(O) as a self-tuning graph toolkit for outcome prediction in event sequence data. The framework is built around T-GCN. Its two level design preserves the semantic distinction between event and sequence information by placing each information type at its corresponding level of the GNN: event attributes are learned through graph convolution over temporally weighted event relations, while sequence attributes are incorporated into the pooled graph representation. O-GCN serves as the one level reference, replicating the same sequence attributes across nodes before graph convolution for direct comparison with the hierarchical organization of T-GCN. TP-GCN and TE-GCN extend T-GCN with a pseudo informative duration embedding and a learned activity embedding, respectively. Each applicable architecture is implemented with GCNConv and GraphConv and optimized through architecture specific automated configuration. The evaluation covers three benchmark datasets with balanced and highly imbalanced outcome distributions, comparisons with attention based GNNs, outcome prediction from incomplete sequence prefixes, and structural ablation analysis.

The results consistently support the design and adaptive configuration of HGCN(O) across the evaluated settings. With automated configuration, the core T-GCN achieves stronger performance than the fixed GAT and TransformerConv baselines, while the close results obtained with GCNConv and GraphConv indicate that this performance is not tied to a particular graph convolution operator. On the imbalanced dataset, the improvement of T-GCN over O-GCN supports the two level organization of event and sequence information, while the behavior of TP-GCN and TE-GCN shows that additional representations can provide further benefit depending on the data and graph operator. The substantial variation in the selected configurations further reflects the need for architecture specific configuration. Performance under partial sequences and the degradation observed after removing temporal edge weights or graph level attributes extend this evidence beyond final classification accuracy. These results show that HGCN(O) benefits from both adaptive configuration and the integrated graph representation itself.

Future work should move HGCN(O) beyond the current benchmark scale and examine its behavior on larger industrial event sequence data, where longer sequences, greater heterogeneity, and more severe class imbalance can place different demands on both graph representation and automated configuration. At larger scale, the two level architecture also provides a common basis for evaluating more expressive graph operators, including attention based models, without changing the underlying treatment of event and sequence information. Such extensions should be accompanied by more cost aware optimization strategies, since the main scalability limitation is likely to shift from sparse graph processing to the computational burden of extensive model search. A practical direction is therefore to balance predictive gain, model complexity, and search cost within the same adaptive framework, particularly for early prediction settings where decisions must be made before complete event sequence information becomes available.

\bibliographystyle{model1-num-names}
\bibliography{ref}

\section*{Statement}
During the preparation of this work, the authors used ChatGPT (OpenAI) for language refinement and proofreading. After using this tool, the authors reviewed and edited the content as needed and take full responsibility for the content of the published article.

\appendix
\renewcommand{\thesection}{Appendix~\Alph{section}}
\setcounter{table}{0} 

\section{Optimized Hyperparameter Configurations for HGCN(O) Models}
\label{sec:appendix}
\begin{sidewaystable}[!htbp]
\centering
\caption{Optimized Configurations of GCNConv and GraphConv Models for Imbalanced Patients Dataset}
\label{app_tab:Imresults}
\scriptsize
\setlength{\tabcolsep}{1pt}

\renewcommand{\arraystretch}{1.05}

\resizebox{\textheight}{!}{%
\begin{tabular}{lccccccccccccccccccccccc}
\toprule
Model   & F1  & B   & G(L) & G(U) & G(A)   & SC  & G(BM)  & G(BE)  & G(D)  & G(M)  & P    & D(L) & D(U) & D(A)   & D(BM)  & D(BE)  & D(D)  & Opt   & LR    & WD    & Sch   & Loss  & L1   \\
\cmidrule[1pt](lr){1-3}
\cmidrule[1pt](lr){4-11}
\cmidrule[1pt](lr){13-18}
\cmidrule[1pt](lr){19-21}
O-G(I)   & 0.859 & 32    & 3     & 224   &  tanh    &  F     &          &          &         &       &  max    & 1     & 200   &  ELU   & 0.6507 & 3.839e-3 & 0.2456 &  RMS  & 1.857e-6 & 6.954e-5 &  Step   & MM    & 4.066e-5 \\
             & (0.0103) &  (142)     &        & 102   &  tanh    &  F     & 0.4504 & 2.13e-3 &         &         &        &        &        &          &          &          &       &  (0.9080,       &         &       & (28,      &        &  \\
             & (0.8738) &       &        & 187   &  GELU    &  T     & 0.2662 & 4.145e-3 &         &         &        &        &        &          &          &          &        & 0.6204,   &   2.888e-8)  &       &  0.4188)     &       &  \\
\midrule
O-C(I)   & 0.8581 & 32    & 3     & 96    &  l\_rl  &  F    & 0.7746 & 7.298e-4 &         &  max    & mean  & 1     & 193   &  ELU   & 0.4741 & 5.322e-4 &          & RMS   & 6.028e-8 & 6.985e-4 &  Exp    &  MM   &  2.013e-5 \\
             & (0.0376) & (77)      &        & 170   &  ReLU       &  T    &             &            &         &  max    &        &        &        &             &             &          &         & (0.9464,         &         & & (0.8956)        &       &  \\
             & (0.8692) &       &        & 123   &  l\_rl &  F    & 0.5253 & 4.217e-4 &         &  mean   &        &        &        &             &             &          &         &  0.6985,        &  9.437e-8)        &          &       &       &  \\
\midrule
T-G(I)    &    0.8639    &    32    &    2  &    88     &    GELU     &F            &           &            &            &           &    max    &   2(S)$^*$    &    69      &    ReLU    &           &           &    0.1601    &    Adam    &    1.248e-3    &    7.736e-4    &    Cos &    CE    &    3.424e-4    \\
                   &   (0.0395)   &   (88)       &           &    151    &  l\_rl    & T          &    0.1764    &    2.471e-3    &            &           &           &       &    133     &    GELU    &   0.9650    &    9.671e-3    &    0.3444    &    (0.8837,    &           &           &   (50, &           &    \\\cmidrule(lr){13-18}
                   &   (0.8762)   &            &           &           &           &            &            &              &            &           &           &   1(C)$^*$    &    188     &    GELU    &    0.2754    &    8.581e-3    &    0.3062    &    0.9405)    &           &           &   3.037e-3)    &           &  \\ 
                   \midrule
T-C(I)   &  0.8599 &  32   &  4     &  148   &  l\_rl  &  T    &  0.9083  &  6.797e-3  &         &  mean   &  max   &  1(S)$^*$     & 53    & GELU  &       &       & 0.2254 &  RMS  &  8.200e-6  &  9.062e-4  &  RP  &  CE   &  2.017e-5  \\ \cmidrule(lr){13-18}
             &  (0.0616)         &   (42)    &        &  82    &  GELU       &  T    &  0.8783  &  6.551e-3  &         &  add    &        & 3(C)$^*$    &  131   &  sp   &  0.4915  &  6.877e-3  &       &  (0.9478,   &       &  (Max,       &  0.2471, &        &  \\
             &  (0.8738)        &       &        &  250   &  ELU        &  T    &  0.6116  &  3.451e-3  &  0.2922  &  add    &        &        & 85    & tanh  &             &          &       &  0.8631, &       &4,          & 9.555e-3,  &       &  \\
             &       &       &        &  54    &  ELU        &  T    &  0.6215  &  2.972e-3  &         &  add    &        &        & 215   &   ReLU         & 0.1363 & 2.086e-3 &       & 3.593e-8 ) &       &          & 8.474e-6)   &   &  \\
\midrule
TP-G(I)   &   0.8595   &   32    &   2(N)$^*$    &   245   &   GELU   &          &         &          &         &          &   add     &  1(S)$^*$   &   228    &  l\_rl  &         &         &         &  Adam  &   1.361e-4   &   8.761e-4  &  OCL  &   CE   &   7.324e-5  \\ \cmidrule(lr){13-18}
              &   (0.0526)  &  (61)       &        &   203   &  GELU   &          &   0.5817  &   9.520e-3  &       &          &           &  2(C)$^*$  &   91     &  l\_rl  &         &         &   0.1195  &  (0.9374)  &          &          & (1.831e-2, &       &  \\\cmidrule(lr){4-11}
              &   (0.8715)  &         & 4(P)  & 38    & GELU  &       & 0.5377 & 2.934e-3 & 0.1011 &          &           &        &  232    &  ELU   &   0.4174  &   9.586e-3  &   0.1100  &   0.9305) &       &          &  0.1203, &       &        \\
              &             &         &        &   194   &  ReLU   &          &          &            &         &          &           &        &         &        &           &            &           &          &           &          &  54000) &       &  \\
              &             &         &        &   196   &  ELU    &          &          &            &         &          &          &        &         &        &           &            &           &          &           &          &           &        &   \\
          &       &       &       &   159   &  sp     &          &   0.2075  &   6.602e-3  &   0.4605  &       &       &       &       &       &       &       &       &       &       &       &       &       &  \\\cmidrule(lr){4-11}
          &       &       & 1(C)$^*$  & 32    & ReLU  & T     &       &       &       &       &       &       &       &       &       &       &       &       &       &       &       &       &  \\
\midrule
TP-C(I)    &    0.8627    &    64    &  2(N)$^*$     &    194    &    tanh    &            &           &            &       & mean  &    add      &   2(S)$^*$    &    243     &   ReLU   &           &           &  0.1235   &   Adam   &    7.155e-4    &    8.389e-4   &   Exp   &    CE    &    1.577e-4   \\
                   &    (0.0564)   &   (86)        &          &    152    &   GELU    &            &    0.2008   &    3.538e-3   &         & add   &             &       &    52      &   ReLU   &    0.5230   &    7.547e-3   &           &   (0.8965,   &            &            &  (0.9602)  &         &   \\\cmidrule(lr){4-11} \cmidrule(lr){13-18} 
                   &    (0.8738)   &           &  1(P)   &  119    &  GELU   &         &         &         &         & add   &             &   2(C)$^*$    &    112     &   ReLU    &    0.6955   &    2.701e-3   &       &    0.9399)  &         &            &         &         &    \\\cmidrule(lr){4-11}
                   &               &           & 3(C)$^*$  &    164    &   GELU    & F     &    0.7264   &    5.268e-3   &           & max   &             &          & 219   & sp    &             &              &    0.2987   &            &             &            &             &         &  \\
                   &               &           &          &    120    &   l\_rl     & F     &    0.2337   &    1.258e-4   &           & max   &            &          &           &          &             &              &             &            &             &            &             &         &  \\
\midrule
TE-G(I)   &    0.8622    &    128    & 5(E)$^*$     & 162   & l\_rl &       & 0.4643 & 2.321e-3 &       &       &    add    &   1(S)$^*$    &    156     &   GELU    &           &           &    0.2540    &    RMS    &    1.425e-5    &    1.567e-4    &    Poly    &    MM    &    2.207e-7   \\\cmidrule(lr){13-18}
                    &    (0.0260)   &   (100)      & (11)$^\dagger$    & 72    & l\_rl &       &       &       &       &       &           &   1(C)$^*$    &    181     &   tanh    &           &           &              &    (0.9008,   &           &           &   (119,   &           &  \\
                    &    (0.8762)   &             &           & 222   & ELU   &       & 0.6657 & 6.918e-3 &       &       &           &           &           &           &           &           &              &          0.2334, &           &           & 1.1527) &       &  \\
                    &               &             &           & 97    & l\_rl &       & 0.7483 & 4.479e-3 & 0.1012 &       &           &           &           &           &           &           &              & 4.253e-8) &           &           &           &       &  \\
                    &               &             &           & 161   & ReLU  &       & 0.3695 & 4.693e-3 &       &       &           &           &           &           &           &           &              &       &           &           &           &       &   \\\cmidrule(lr){4-11} 
          &       &       & 5(N)$^*$  &    147    &   tanh    &            &            &            &  0.4960   &           &       &       &       &       &       &       &       &       &       &       &       &       &  \\
          &       &       &       &    241    &  tanh     &            &    0.6208    &    2.883e-3    &  0.3867   &           &       &       &       &       &       &       &       &       &       &       &       &       &  \\
          &       &       &       &     57     &  l\_rl     &            &    0.0287    &    6.263e-4    &           &           &       &       &       &       &       &       &       &       &       &       &       &       &  \\
          &       &       &       &    162    &  tanh     &            &             &              &           &           &       &       &       &       &       &       &       &       &       &       &       &       &  \\
          &       &       &       &    184     &  l\_rl     &            &             &              &           &           &       &       &       &       &       &       &       &       &       &       &       &       &     \\\cmidrule(lr){4-11}
          &       &       & 1(C)$^*$  & 213   &  l\_rl& T     &       &       &       &       &       &       &       &       &       &       &       &       &       &       &       &       &  \\ 
\midrule
TE-C(I)    &     0.8662     &     64     &  1(E)$^*$     &  199    &  GELU    &   &0.8902   &  4.456e-3         &  0.1767    &  max    &  mean     &    1(S)$^*$     &     117      &    sp &             &             &             &     Adam     &     2.589e-3     &     5.502e-4     &     Cos     &     CE     &     1.791e-4  \\ \cmidrule(lr){4-11}\cmidrule(lr){13-18} 
                        &     (0.0467)    &    (98)     &  (12)$^\dagger$    &  222    &  l\_rl   &             &            &         &           &  mean     &            & 2(C)$^*$  &     119      &    sp &             &             &             &     (0.9420,    &             &             &    (83, &             &  \\
                        &     (0.8785)    &             & 2(N)$^*$  &  128    &  l\_rl   &             &            &         &  0.2398   &  max      &            &       &     89       &    l\_rl     &             &             &             &           0.9709) &             &             &  2.605e-3)  &             &    \\\cmidrule(lr){4-11} 
                        &                 &             & 1(C)$^*$  &  130    &  l\_rl   &  T          &  0.6489   &  9.365e-3  &          &  max      &            &             &             &             &             &             &             &         &             &             &             &             &    \\\bottomrule
\end{tabular}%
}
\parbox{\textheight}{%
\scriptsize
\textbullet\ Model: G: GCNConv; C: GraphConv
\textbullet\ A: Accuracy; F1: F1(loss std)(Acc); B: Batch size (Best Epoch); G(L)/D(L): Number of hidden G(GCN)/D(Dense) layers; G(U)/D(U): Units; G(A)/D(A): Activation functions; G(BE)/D(BE): Batch normalization epsilon; G(BM)/D(BM): Batch normalization momentum;  G(D)/D(D): Dropout rates; SC: Skip Connection flag; G(M): Aggregation method for GraphConv; P: Pooling Method; Opt: Optimizer; LR: Learning Rate; WD: Weight Decay; Sch: Learning Rate Scheduler; Loss: Loss function; L1: L1 regularization; Empty cell in (BM)/(BE)/(D): No batch normalization or dropout applied. 
\textbullet\ $*$:(N): Node input layer (E): Embedding layer; (P): Pseudo-embedding input layer; (S): Graph level input layer; (C): Concatenation GCN/Dense layer; $\dagger$: Embedding dimensions
\textbullet\ l\_rl: LeakyReLU; sp: Softplus; CE: CrossEntropy; MM: MultiMargin.
\textbullet\ Adam: Adam ($\beta_1$, $\beta_2$); SGD: SGD(momentum); RMS:RMSprop($\alpha$, momentum, eps); Step: Step(step size, $\gamma$); Exp: Exponential($\gamma$); RP: Reduce-on-Plateau(mode, factor, patience, threshold, eps); Poly: Polynomial(total\_iters, power); Cos: Cos(T\_max, eta\_min); OCL: One Cycle(max, pct\_start, total\_steps); Cy: Cyclic(base, max, step\_size\_up)
}
\end{sidewaystable}

%========================================================
% BPI12-A
%========================================================
\begin{sidewaystable}[!htbp]
\centering
\caption{Optimized Configurations of GCNConv and GraphConv Models for BPI12-A}
\label{app_tab:BPIResultsA}
\scriptsize
\setlength{\tabcolsep}{1pt}
\renewcommand{\arraystretch}{1.05}

\resizebox{\textheight}{!}{%
\begin{tabular}{lccccccccccccccccccccccc}
\toprule
Model & A & B & G(L) & G(U) & G(A) & SC & G(BM) & G(BE) & G(D) & G(M) & P & D(L) & D(U) & D(A) & D(BM) & D(BE) & D(D) & Opt & LR & WD & Sch & Loss & L1 \\
\cmidrule(lr){1-3}
\cmidrule(lr){4-11}
\cmidrule(lr){13-18}
\cmidrule(lr){19-21}

O-G(A)  & 1 & 32 & 2 & 188 & ELU  & T & 0.4679 & 6.673e-3 & 0.3680 &       & max  & 3 & 171 & ReLU & 0.4045 & 1.212e-3 & 0.2913 & Adam & 4.236e-2 & 6.323e-3 & Exp  & MM & 4.557e-4 \\
        &   & (18) &   & 85  & ELU  & T & 0.6117 & 5.621e-3 & 0.4486 &       &      &   & 109 & ELU  & 0.1688 & 3.723e-3 & 0.4262 & (0.8863, &       &       & (0.9441) &    &    \\
        &   &      &   &     &      &   &        &          &        &       &      &   & 182 & ELU  &        &          &        & 0.9120)  &       &       &          &    &    \\
\midrule

O-C(A)  & 1 & 32 & 3 & 211 & ReLU & F &        &          & 0.1597 & add   & mean & 1 & 189 & ReLU & 0.6328 & 7.977e-3 & 0.1636 & Adam & 9.233e-4 & 6.863e-3 & Poly & MM & 9.753e-5 \\
        &   & (21) &   & 100 & sp   & F & 0.0507 & 6.398e-4 & 0.1852 & add   &      &   &     &      &        &          &        & (0.8887, &       &       & (129,    &    &    \\
        &   &      &   & 75  & ELU  & F &        &          & 0.4934 & add   &      &   &     &      &        &          &        & 0.9167)  &       &       & 0.5612)  &    &    \\
\midrule

T-G(A)  & 1 & 64 & 1 & 169 & GELU & T & 0.6317 & 2.408e-3 &        &       & max  & 2(S)$^*$ & 41  & l\_rl &        &          &        & RMS  & 4.547e-3 & 8.561e-3 & Cos  & MM & 5.908e-4 \\
        &   & (29) &   &     &      &   &        &          &        &       &      &          & 105 & l\_rl &        &          &        & (0.9154, &       &       & (51,     &    &    \\
        &   &      &   &     &      &   &        &          &        &       &      & 1(C)$^*$ & 34  & l\_rl &        &          &        & 0.8371,  & 9.053e-8 &   & 6.309e-4) & & \\
\midrule

T-C(A)  & 1 & 32 & 2 & 151 & tanh & F & 0.2745 & 1.149e-3 & 0.4690 & mean  & max  & 1(S)$^*$ & 111 & tanh  &        &          &        & RMS  & 7.234e-4 & 4.415e-3 & Cy   & MM & 5.292e-4 \\
        &   & (25) &   & 128 & sp   & F & 0.4512 & 6.046e-3 &        & max   &      & 1(C)$^*$ & 169 & l\_rl &        &          & 0.243  & (0.9111, & 0.7603, & 8.523e-8) & (0.0019, & 5.0392e-2, & 25) \\
\midrule

TE-G(A) & 1 & 64 & 2(E)$^*$ & 48  & l\_rl &   &        &          &        &      & add  & 2(S)$^*$ & 48  & GELU &        &          & 0.3663 & Adam & 2.393e-3 & 2.099e-3 & RP   & CE & 3.930e-4 \\
        &   & (31) & (16)$^\dagger$ & 237 & ELU  &   & 0.6045 & 8.165e-3 & 0.2205 &      &      &          & 128 & GELU &        &          &        & (0.9692, &       &       & (Max,    &    &    \\
        &   &      & 5(N)$^*$ & 128 & ReLU &   & 0.1164 & 7.970e-3 &        &      &      & 1(C)$^*$ & 83  & ReLU &        &          & 0.3612 & 0.9394)  &       &       & 0.8104,  &    &    \\
        &   &      &          & 192 & GELU &   & 0.5109 & 8.158e-3 &        &      &      &          &     &      &        &          &        &          &       &       & 23,      &    &    \\
        &   &      &          & 228 & l\_rl &  & 0.1853 & 8.930e-3 &        &      &      &          &     &      &        &          &        &          &       &       & 7.53e-3, &    &    \\
        &   &      &          & 201 & ReLU &   &        &          &        &      &      &          &     &      &        &          &        &          &       &       & 1.620e-5)&    &    \\
        &   &      &          & 116 & ELU  &   &        &          &        &      &      &          &     &      &        &          &        &          &       &       &          &    &    \\
        &   &      & 1(C)$^*$ & 128 & GELU & T &        &          & 0.2656 &      &      &          &     &      &        &          &        &          &       &       &          &    &    \\
\midrule

TE-C(A) & 1 & 128 & 4(E)$^*$ & 235 & ELU  &   &        &          &        & add  & mean & 3(S)$^*$ & 131 & ReLU &        &          & 0.1011 & Adam & 4.579e-4 & 2.018e-3 & RP   & MM & 3.315e-4 \\
        &   & (23) & (18)$^\dagger$ & 111 & l\_rl & &      &          & 0.2249 & max  &      &          & 215 & GELU &        &          &        & (0.9477, &       &       & (Max,    &    &    \\
        &   &      &          & 230 & l\_rl &   &        &          &        & mean &      &          & 194 & tanh & 0.3789 & 8.481e-3 &        & 0.9033)  &       &       & 0.8993,  &    &    \\
        &   &      &          & 105 & GELU &   & 0.5153 & 6.239e-3 & 0.1027 & max  &      & 1(C)$^*$ & 156 & ReLU & 0.5428 & 9.075e-3 &        &          &       &       & 1,       &    &    \\
        &   &      & 4(N)$^*$ & 123 & GELU &   & 0.3155 & 6.208e-3 & 0.2360 & mean &      &          &     &      &        &          &        &          &       &       & 1.042e-4,&    &    \\
        &   &      &          & 62  & l\_rl &  &        &          &        & max  &      &          &     &      &        &          &        &          &       &       & 9.602e-5)&    &    \\
        &   &      &          & 92  & GELU &   &        &          & 0.2555 & mean &      &          &     &      &        &          &        &          &       &       &          &    &    \\
        &   &      &          & 119 & GELU &   & 0.7320 & 7.155e-3 &        & mean &      &          &     &      &        &          &        &          &       &       &          &    &    \\
        &   &      & 1(C)$^*$ & 59  & tanh & T & 0.3805 & 4.642e-3 &        & max  &      &          &     &      &        &          &        &          &       &       &          &    &    \\
\bottomrule
\end{tabular}%
}

\parbox{\textheight}{%
\scriptsize
\textbullet\ Abbreviations are the same as Table~\ref{app_tab:Imresults}.}
\end{sidewaystable}

%========================================================
% BPI12-O
%========================================================
\begin{sidewaystable}[!htbp]
\centering
\caption{Optimized Configurations of GCNConv and GraphConv Models for BPI12-O}
\label{app_tab:BPIResultsO}
\scriptsize
\setlength{\tabcolsep}{1pt}
\renewcommand{\arraystretch}{1.05}

\resizebox{\textheight}{!}{%
\begin{tabular}{lccccccccccccccccccccccc}
\toprule
Model & A & B & G(L) & G(U) & G(A) & SC & G(BM) & G(BE) & G(D) & G(M) & P & D(L) & D(U) & D(A) & D(BM) & D(BE) & D(D) & Opt & LR & WD & Sch & Loss & L1 \\
\cmidrule(lr){1-3}
\cmidrule(lr){4-11}
\cmidrule(lr){13-18}
\cmidrule(lr){19-21}

O-G(O)  & 1 & 32 & 2 & 158 & tanh & T & 0.0281 & 2.547e-4 & 0.4896 &     & add  & 2 & 130 & l\_rl & 0.4479 & 9.357e-3 & 0.3127 & SGD  & 1.964e-3 & 3.927e-3 & Cy   & MM & 6.112e-4 \\
        &   & (13) &   & 216 & ELU  & T & 0.9216 & 5.337e-3 & 0.3833 &     &      &   & 180 & l\_rl & 0.3666 & 9.1e-4   &        & (0.0868) & (7.214e-3, & 9.382e-2, & 36) & & \\
\midrule

O-C(O)  & 1 & 32 & 5 & 111 & tanh & F & 0.5339 & 9.711e-3 & 0.3955 & max & max  & 1 & 210 & sp    & 0.3473 & 2.428e-3 & 0.3955 & Adam & 2.223e-3 & 3.593e-3 & Step & MM & 7.391e-5 \\
        &   & (20) &   & 128 & sp   & T &        &          &        & add &      &   &     &       &        &          &        & (0.9081, &       &       & (20      &    &    \\
        &   &      &   & 124 & ReLU & F &        &          &        & add &      &   &     &       &        &          &        & 0.9283)  &       &       & 0.7002)  &    &    \\
        &   &      &   & 44  & GELU & F &        &          & 0.3403 & max &      &   &     &       &        &          &        &          &       &       &          &    &    \\
        &   &      &   & 71  & l\_rl& T &        &          &        & mean&      &   &     &       &        &          &        &          &       &       &          &    &    \\
\midrule

T-G(O)  & 1 & 512 & 3 & 154 & sp   & F & 0.6040 & 5.235e-3 & 0.1950 &     & max  & 3(S)$^*$ & 156 & sp   & 0.6054 & 8.760e-3 & 0.4070 & Adam & 3.951e-4 & 7.458e-3 & RP   & MM & 9.109e-4 \\
        &   & (16) &   & 238 & ReLU & T &        &          & 0.4380 &     &      &          & 53  & ReLU & 0.7485 & 2.799e-3 & 0.2210 & (0.975, &       & (Max,    & 0.5128, & & \\
        &   &      &   & 163 & l\_rl& F & 0.0144 & 9.534e-3 &        &     &      &          & 111 & ELU  & 0.3305 & 2.911e-3 &        & 0.926)  &       & 16,      & 4.604e-4& & \\
        &   &      &   &     &      &   &        &          &        &     &      & 1(C)$^*$ & 120 & ReLU & 0.3416 & 2.631e-3 & 0.3160 &         &       &          & 7.449e-5& & \\
\midrule

T-C(O)  & 1 & 64 & 5 & 197 & ReLU & F &        &          & 0.2854 & add & add  & 3(S) & 247 & ReLU & 0.0678 & 2.667e-3 &        & Adam & 9.804e-3 & 5.032e-3 & OCL  & MM & 1.449e-5 \\
        &   & (25) &   & 98  & ELU  & F & 0.3419 & 5.304e-3 &        & add &      &      & 124 & l\_rl & 0.8186 & 6.601e-3 & 0.1938 & (0.8851, &       &       & (0.0879, & & \\
        &   &      &   & 162 & l\_rl& F &        &          &        & max &      &      &     &       &        &          &        & 0.9008)  &       &       & 0.449)   & & \\
        &   &      &   & 172 & ReLU & F &        &          &        & max &      & 1(C) & 227 & ELU  &        &          &        &          &       &       &          & & \\
        &   &      &   & 191 & GELU & F & 0.3995 & 3.511e-3 & 0.4485 & max &      &      &     &      &        &          &        &          &       &       &          & & \\
\midrule

TE-G(O) & 1 & 64 & 1(E)$^*$ & 125 & GELU &   & 0.2835 & 8.908e-3 & 0.2889 &      & max  & 3(S)$^*$ & 103 & sp   & 0.6693 & 3.278e-3 &        & RMS  & 1.907e-4 & 8.362e-3 & Poly & MM & 8.788e-4 \\
        &   & (17) & (13)$^\dagger$ & 50  & ReLU &   &        &          &        &      &      &          & 93  & GELU &        &          &        & (0.9021, &       & (93,     &          &    &    \\
        &   &      & 4(N)$^*$ & 180 & ReLU &   & 0.0103 & 2.119e-3 &        &      &      &          & 95  & ELU  &        &          &        & 0.7441,  &       & 1.1485)  &          &    &    \\
        &   &      &          & 164 & GELU &   &        &          & 0.2460 &      &      & 1(C)$^*$ & 114 & ELU  &        &          & 0.4481 & 9.331e-8)&       &          &          &    &    \\
        &   &      &          & 66  & l\_rl&  & 0.3423 & 6.425e-3 & 0.1233 &      &      &          &     &      &        &          &        &          &       &          &          &    &    \\
        &   &      & 4(C)$^*$ & 145 & GELU & T & 0.7255 & 8.292e-3 & 0.1577 &      &      &          &     &      &        &          &        &          &       &          &          &    &    \\
        &   &      &          & 114 & tanh & T &        &          &        &      &      &          &     &      &        &          &        &          &       &          &          &    &    \\
        &   &      &          & 202 & ReLU & T & 0.6147 & 5.549e-3 & 0.2826 &      &      &          &     &      &        &          &        &          &       &          &          &    &    \\
        &   &      &          & 73  & l\_rl& T & 0.2547 & 6.275e-3 & 0.2660 &      &      &          &     &      &        &          &        &          &       &          &          &    &    \\
\midrule

TE-C(O) & 1 & 32 & 5(E)$^*$ & 34  & sp   &   & 0.3921 & 6.222e-3 &        & mean & mean & 2(S)$^*$ & 130 & ELU  &        &          &        & SGD  & 1.165e-3 & 7.960e-3 & OCL  & MM & 8.182e-4 \\
        &   & (31) & (21)$^\dagger$ & 192 & l\_rl&   & 0.6985 & 2.935e-3 & 0.2289 & max  &      &          & 156 & ELU  &        &          &        & (0.6997) &       &          & (2.912e-2, & & \\
        &   &      &          & 48  & l\_rl&   &        &          &        & add  &      & 2(C)$^*$ & 204 & ReLU &        &          &        &          &       &          & 0.2517,    & & \\
        &   &      &          & 111 & l\_rl&   &        &          &        & add  &      &          & 246 & sp   & 0.2550 & 7.048e-3 & 0.3087 &          &       &          & 61,000)    & & \\
        &   &      &          & 232 & l\_rl&   & 0.0729 & 9.9516e-4& 0.1193 & mean &      &          &     &      &        &          &        &          &       &          &            & & \\
        &   &      & 4(N)$^*$ & 176 & GELU &   & 0.3532 & 9.875e-3 & 0.2951 & mean &      &          &     &      &        &          &        &          &       &          &            & & \\
        &   &      &          & 98  & ELU  &   &        &          &        & mean &      &          &     &      &        &          &        &          &       &          &            & & \\
        &   &      &          & 243 & sp   &   &        &          &        & add  &      &          &     &      &        &          &        &          &       &          &            & & \\
        &   &      &          & 213 & ELU  &   & 0.0218 & 9.528e-3 &        & max  &      &          &     &      &        &          &        &          &       &          &            & & \\
        &   &      & 1(C)$^*$ & 207 & sp   & F &        &          &        & add  &      &          &     &      &        &          &        &          &       &          &            & & \\
\bottomrule
\end{tabular}%
}

\parbox{\textheight}{%
\scriptsize
\textbullet\ Abbreviations are the same as Table~\ref{app_tab:Imresults}.}
\end{sidewaystable}

\vfill

\end{document}